\documentclass[final,5p,times,twocolumn]{elsarticle}

\usepackage{amssymb}
\usepackage{amsmath}

\newcommand*{\method}{IDEAL}
\usepackage{bbm}
\usepackage{multirow}
\usepackage{makecell}
\usepackage{xcolor}
\usepackage{booktabs}
\usepackage{hyperref}
\usepackage{subcaption} 
\usepackage{fontawesome} 
\definecolor{second}{HTML}{FFDAB9}
\definecolor{best}{HTML}{FFC1C1}
\usepackage{colortbl}

\journal{KNOWLEDGE-BASED SYSTEMS}

\begin{document}

\begin{frontmatter}



\author{Jie Cao  \fnref{label2}}
\ead{caojie@zju.edu.cn}

\author{Dian Jiao \fnref{label2}}
\ead{jd_dcd@zju.edu.cn}

\author{Yang Dai \fnref{label2}}
\ead{yangdai@zju.edu.cn}

\author{Rolan Yan \fnref{label3}}
\ead{rolanyan@tencent.com}

\author{Wenqiao Zhang \corref{cor1} \fnref{label2}}
\ead{wenqiaozhang@zju.edu.cn}

\author{Siliang Tang \fnref{label2}}
\ead{siliang@zju.edu.cn}

\cortext[cor1]{Corresponding author}
\fntext[label2]{Zhejiang University}
\fntext[label3]{Tencent, Wechat}

\title{IDEAL: Leveraging Infinite and Dynamic Characterizations of Large Language Models for Query-focused Summarization} 


\author{} 


\begin{abstract}
Query-focused summarization (QFS) aims to produce summaries that answer particular questions of interest, enabling greater user control and personalization. The advent of large language models (LLMs), shows their impressive capability of textual understanding through large-scale pretraining, which implies the great potential of extractive snippet generation. In this paper, we systematically investigated two indispensable characteristics that the LLMs-based QFS models should be harnessed, \emph{Efficiently Fine-grained Query-LLM Alignment} and \emph{Lengthy Document Summarization}, respectively. Correspondingly, we propose two modules called Query-aware HyperExpert and Query-focused Infini-attention to access the aforementioned characteristics. These innovations pave the way for broader application and accessibility in the field of QFS technology. Extensive experiments conducted on existing QFS benchmarks indicate the effectiveness and generalizability of the proposed approach. 
\end{abstract}




\end{frontmatter}


\section{Introduction}

In today’s world, where we are constantly bombarded with vast amounts of text, the ability to efficiently summarize information has become crucial.
Textual summarization~\citep{gambhir2017recent} is the process of condensing a lengthy document into a succinct and digestible version while preserving the most crucial information, enabling quicker understanding and better management of information.
As everyone has unique needs and interests in real-life scenarios, necessitating summarizers that succinctly address the information needed for a specific query by extracting
essential information from documents, \emph{i.e.}, \textbf{Q}uery-\textbf{F}ocused \textbf{S}ummarization (QFS)~\citep{daume2009bayesian}. This task involves analyzing the content to identify key themes and then highlighting these in the summary, which attracts increasing attention in the textual summarization community.

Traditionally, QFS has used extract-then-summarize methods~\citep{zhong_qmsum_2021,wang_squality_2022,amar_openasp_2023} that rely on the most relevant spans of text from a candidate document based on the prevalence of query terms. However, real-world QFS tasks require a comprehensive and in-depth understanding of complex and lengthy documents to generate high-quality, relevant summaries. Further onwards, the triumph of Large Language Models (LLMs) such as the GPT series \citep{achiam2023gpt}, LLaMA \citep{touvron2023llama}, and other open-source LLMs showcased the power of large-scale pretraining in understanding, reasoning, and generating intricate textual patterns. In addition, their remarkable effectiveness in downstream applications~\citep{yang2025power,cary2024herding} further highlights their potential, opening up new opportunities for QFS. 

However, there has been relatively little investigation into LLM-based QFS methods \citep{yang_exploring_2023}. 
Our primary goal in this paper is to close this gap correspondingly by proposing two indispensable characteristics that should be harnessed by LLMs while dealing with QFS:
(i) \textbf{Efficiently Fine-grained Query-LLM Alignment}, 
as commonly known, the pre-trained LLMs are powerful when transferred to downstream tasks with instruction tuning \citep{ouyang_instructGPT_2022}, this also applies to the QFS task when the LLMs specialized for user's interests.
However, as the parameter number grows exponentially to billions or even trillions, training the fully fine-tuned model for each downstream task becomes very inefficient. 
Moreover, the simple approach of concatenating the query to the input document proves insufficient for effectively guiding the model to focus on the query while generating the summary. Due to the small proportion of the query length in the overall input (e.g., in the QMSum \citep{zhong_qmsum_2021} dataset, the average token counts for queries and documents are 15 and 13,227, respectively), the query's control over the model tends to be relatively weak in attention-based models during summary generation.
Therefore, the foundation of more effective LLM-based QFS lies in achieving fine-grained query-LLM alignment through efficient learning.
 (ii) \textbf{Lengthy Document Summarization}, QFS tasks usually involve long documents. However, self-attention-based LLMs have been shown to struggle with handling such long text inputs due to the quadratic complexity of the attention mechanism in terms of both memory usage and computation time. How to process lengthy documents under limited memory is also an important characteristic of LLM-based QFS approaches.
Summing up, these characteristics necessitate a thorough reevaluation of QFS and its corresponding solutions with LLMs.

Based on the aforementioned insights, we propose \textbf{I}nfinite and \textbf{D}ynamic larg\textbf{E} langu\textbf{A}ge mode\textbf{L}-based framework, abbreviated as \method{} \footnote{Code:\url{https://github.com/DCDmllm/IDEAL-Summary}} 
for ideal QFS, which consists of two modules: \textbf{Query-aware HyperExpert} and \textbf{Query-focused Infini-attention}, achieving the two indispensable characteristics, respectively. 

\begin{figure*}[t]
    \centering
    \includegraphics[width=1\textwidth, trim=0.05in 0.05in 0.05in 0.05in, clip]{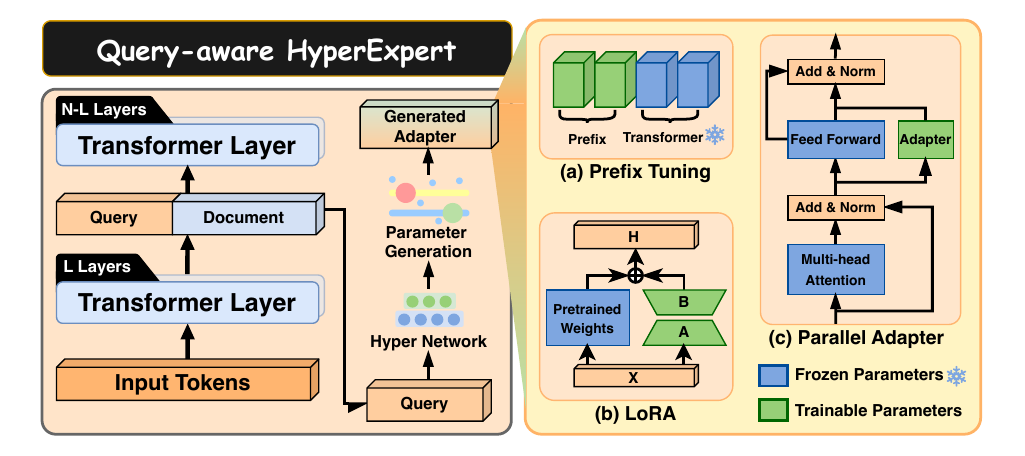}
    \caption{ Overview of \method{}. We place a regular (non-generated) PEFT Adapter layer in the first $l$ layers, and then use the hidden states of the query instruction to generate the Adapter's parameters of the last $N$-$l$ layers.}
    \label{fig:framework}
    \vspace{-3mm}
\end{figure*}

The Query-aware HyperExpert (Figure~\ref{fig:framework}) leverages the parameter-efficient fine-tuning (PEFT) \citep{peft} strategies that enable a pre-trained LLM to perform a new QFS task with minimal parameter updates. Innovatively, we tailor the previous PEFT approaches to QFS tasks with a HyperNetwork~\cite{ha2016hypernetworks}, which can dynamically generate the strongly correlated instance-level PEFT Adapter's parameters according to users' queries. Such dynamic characterization allows us to achieve the best of both worlds by adjusting the LLM's parameters while encouraging the model to adapt to each instance. By doing so, efficient and fine-grained query-LLM alignment can be achieved.
Notably, we develop three types of HyperExpert, include \method{}$_{Prompt}$, \method{}$_{PAdapter}$, and \method{}$_{LoRA}$ based on Prompt-tuning \citep{lester2021prompt-tuning}, Parallel Adapter \citep{he2022parallel_adapters}, and Low-Rank Adaptation (LoRA) \citep{hu2021lora} respectively. 

\begin{figure}[ht]
    \centering
    \includegraphics[width=0.5\textwidth, trim=0.05in 0.05in 0.05in 0.05in, clip]{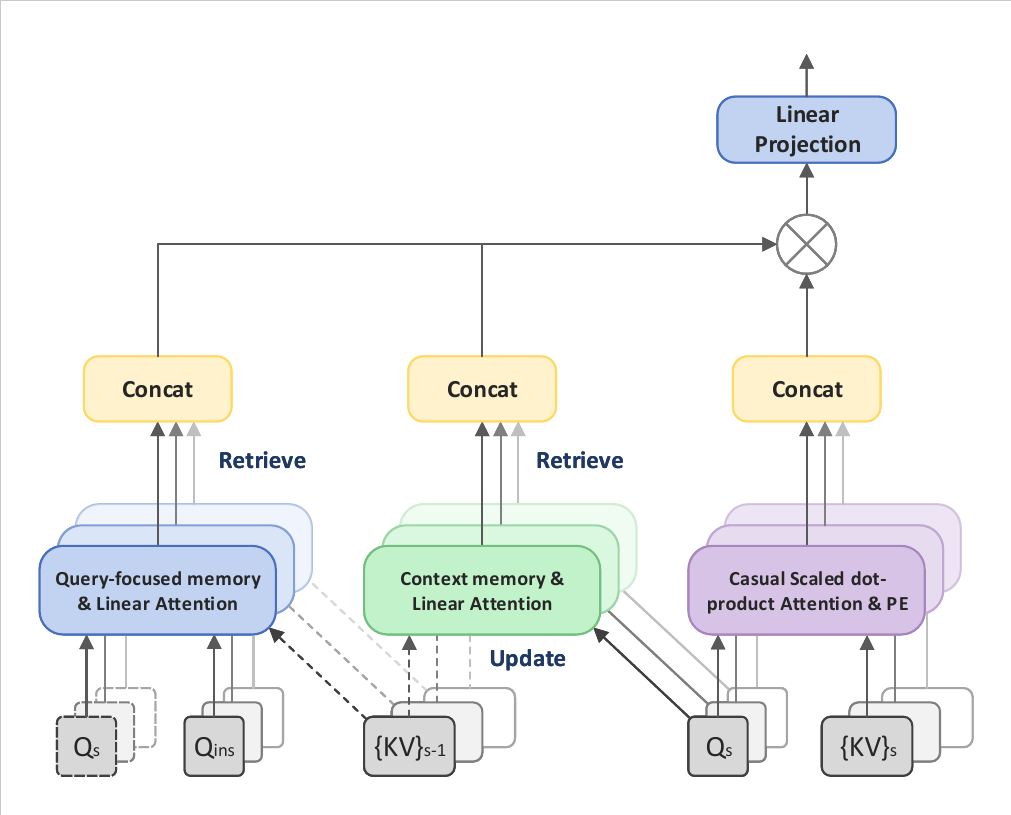}
    \caption{Query-focused Infini-attention has a long-term context memory and a query-focused memory with linear attention for processing infinitely long contexts. $KV_{s-1}$ and $KV_s$ are attention keys and values for previous and current input segments, respectively. $Q$ represents the attention queries for the current input segment, while $Q_{ins}$ refers to the attention queries for the input query instruction. PE signifies position embeddings. }
    \label{fig:QF-INF}
\end{figure}

To enable Transformer-based LLMs to handle extremely long inputs for QFS tasks under limited memory constraints, we propose a Query-focused Infini-Attention (Figure~\ref{fig:QF-INF}) module that can be seamlessly integrated into the Query-aware HyperExpert framework. The Query-focused Infini-Attention builds upon the Infini-Attention mechanism \citep{munkhdalai2024infinitransformer}, which enhances the standard Transformer architecture by introducing compressive memory and a long-term linear attention mechanism. Specifically designed for QFS tasks, the Query-focused Infini-Attention incorporates a Query-focused memory block to preserve critical query-related document details, effectively mitigating the loss of essential information during the compression of query instructions and extremely long input documents. 

Our contributions can be summarized as follows:

\begin{itemize}
  \item We explored query-focused PEFT methods and proposed a method, \method{}, that tunes instance-level PEFT approaches according to query instructions, enhancing the model's fine-grained instruction-following capabilities.
  \item We propose to incorporate a query-focused infini-attention module to process long text under low memory resources for QFS tasks. For example, \method{} with the backbone model LLAMA2-7B can process datasets where the average length of \textbf{input tokens is 13,000 on a single 24GB Nvidia GeForce RTX 3090}.
  \item We performed extensive and rigorous experiments across multiple QFS datasets. \method{} significantly outperforms other baselines. 
\end{itemize}

\section{Related Works}
\subsection{Query-focused Summarization}
\citet{tan_summarizing_any_2020} and \citet{yang_oasum_2023} address QFS by prepending the query or aspect to the input document and fine-tuning pre-trained models in an end-to-end manner. 
\citet{zhong_qmsum_2021}, \citet{wang_squality_2022}, and \citet{amar_openasp_2023} employ extract-then-summarize strategies that use a filter model to extract key parts of the document based on the query, then fitting the shorter text into a summarizer. 
\citet{vig2022exploringNeuQF} use an encoder to compute the local attention of a segmented document. The resulting encodings are then concatenated into a single embedding sequence and passed to a decoder model to generate the summary. \citet{pagnoni2023socratic} introduces a question-driven, unsupervised pre-training objective, specifically designed to improve controllability in summarization tasks. \citet{Sotudeh2023QontSumOC} propose a contrastive learning method aimed at improving the relevance of summaries to a given query.
\citet{yang_exploring_2023} reveal that the performance of ChatGPT is comparable to traditional fine-tuning methods in terms of ROUGE scores on QFS tasks.
\subsection{Long-context Transformers}
LED \citep{beltagy2020longformer} employs a more efficient self-attention pattern that allows the model to scale to long documents. Unlimiformer \citep{bertsch2024unlimiformer} enhances pre-trained models like BART \citep{lewis_bart_2019} to handle unlimited inputs without additional learned weights by employing a retrieval-based long-context method. Infini-transformer \citep{munkhdalai2024infinitransformer} integrates long-term context compressive memory into vanilla transformers, enabling Transformer-based LLMs to scale to infinitely long contexts after full continual pre-training. Unlike the Infini-transformer, we explore the compressive memory method on adapter-based PEFT of LLMs and design a query-focused Infini-attention for QFS tasks.
\subsection{Hypernetwork-based Methods}
\citet{ivison_hyperdecoders_2022} investigates input-conditioned hypernetworks for multi-tasking in NLP, which generate parameter-efficient adaptations for a decoder using a hypernetwork conditioned on the output of an encoder. \citet{he_hyperprompt_2022} incorporates hypernetworks into prompt-based, task-conditioning Transformer models in a multi-task setting, allowing the network to learn task-specific feature maps. \citet{zhang2024hyperllava} employs hypernetworks to generate adaptive parameter shifts for a visual projector and an LLM in multimodal tasks.

\section{Methodology}

\subsection{Query-aware HyperExpert Module}\label{sec:qf-hyper-adapters}
Given a dataset with input text pairs containing a query and a document, outputs in the form of a summary, and a pre-trained LLaMA with an $N$-layer transformer, \method{} used three kinds of PEFT adapters to fine-tune LLaMA to generate query-focused summaries respectively. For example, \method{}$_{LoRA}$, we place a regular (non-generated) LoRA module in the first $l$ layers, then we use the hidden representation $\boldsymbol{H}^l_{query}$ of query in $l$-th layer as the input of a Hypernetwork to generate the LoRA parameters for the last $N-l$ layers. 

\subsubsection{PEFT approaches}

\paragraph{Prompt tuning}
As shown in Figure~\ref{fig:framework}(a), Prompt tuning can add soft prompts to the hidden states in attention layers to guide model learning and adapt to new tasks, where only the soft prompts are updated during training. LLaMA-Adapter-v1 \citep{zhang_llama-adapter_2023} introduces a zero-initialized attention mechanism into prompt tuning, which adaptively incorporates the knowledge from soft prompts. We use this LLaMA-Adapter-v1 as our prompt tuning baseline. In each transformer layer, let $\boldsymbol{P} \in \mathbb{R}^{K \times d}$ denote the learnable prompts, the zero-initialized prompt tuning in the attention block computed as:
\begin{equation}
    \boldsymbol{Q} = \boldsymbol{xW}_q ,
\end{equation}
\begin{equation} \label{eq:prompt}
    \boldsymbol{h} = Attn(\boldsymbol{Q}, \boldsymbol{xW}_k, \boldsymbol{xW}_v) +
    g \cdot Attn(\boldsymbol{Q} , \boldsymbol{PW}_k, \boldsymbol{P}\boldsymbol{W}_{v}),
\end{equation}
where $g$ is a learnable scalar for each head initialized with zero, $\boldsymbol{W}_q, \boldsymbol{W}_k, \boldsymbol{W}_v$ are frozen pre-trained matrices. 

\paragraph{Parallel adapter}
Parallel adapters \citep{he2022parallel_adapters} aim to incorporate additional learnable networks in parallel with distinct sublayers in the backbone model. To reduce the number of parameters, small bottleneck networks are used as parallel adapters: a down-projection with $\boldsymbol{W}_{down} \in \mathbb{R}^{d \times r}$ to project the input $\boldsymbol{x}$ to a lower-dimensional space specified by bottleneck dimension $r$, followed by a nonlinear activation function $f(\cdot)$, an up-projection with $\boldsymbol{W}_{up} \in \mathbb{R}^{r \times d}$ to project back to the input size. The parallel adapter can be defined as:
\begin{equation}
    \boldsymbol{h} = F(\boldsymbol{x}) + f(\boldsymbol{xW}_{down})\boldsymbol{W}_{up},
\end{equation}
where $F(\cdot)$ represents the feedforward layer or self-attention modules here.

In transformer-based LLMs, parallel adapters can be applied to both the feedforward and self-attention modules in each transformer block. \citet{hu_llm-adapters_2023} conducted experiments showing that applying parallel adapters only to the feedforward module achieves the best results on math reasoning datasets. As shown in Figure~\ref{fig:framework}(c), we also apply parallel adapters only to feedforward modules in LLaMA's transformer block. 

\paragraph{LoRA}
LoRA \citep{hu2021lora} adds trainable low-rank decomposition matrices in parallel to existing weight matrices (Figure~\ref{fig:framework}(b)). For a pre-trained weight matrix $\boldsymbol{W}\in \mathbb{R}^{d\times k}$, LoRA constrains its update by adding low-rank matrix pairs, resulting in $\boldsymbol{W} + \Delta\boldsymbol{W} = \boldsymbol{W} + \boldsymbol{BA}$, where $\boldsymbol{B}\in \mathbb{R}^{d\times r}$, $\boldsymbol{A}\in \mathbb{R}^{r\times k}$, and the rank $r \ll \min(d,k)$. During training, $\boldsymbol{W}$ is frozen while $\boldsymbol{B}$ and $\boldsymbol{A}$ are trainable. LoRA initializes $\boldsymbol{A}$ randomly and $\boldsymbol{B}$ to zero, ensuring that $\Delta\boldsymbol{W} = \boldsymbol{BA}$ starts from zero at the beginning of training, thereby preserving the pre-trained knowledge as much as possible.

\subsubsection{Adapter-based HyperExpert}
Previous works \citep{ivison_hyperdecoders_2022,zhao_hypermoe_2024} indicate that hypernetworks can learn the parameter information of the main neural network under different input scenarios and efficiently adjust the target network's parameters to adapt to this information. We propose leveraging a Hypernetwork to generate adapters conditioned on query instructions, enhancing the model's query-focused capabilities.

Our HyperExpert is a Hypernetwork that consists of an \textbf{encoder} that transforms the mean-pooling of the query representation $\boldsymbol{H}_{query}$ into a low-dimensional representation $\boldsymbol{h} \in \mathbb{R}^{b}$, and a \textbf{decoder} that converts $\boldsymbol{h}$ into the parameters of the target PEFT adapters. 
The \textbf{encoder} is consistent across all three types of HyperExpert and is computed as follows: 
\begin{equation} \label{eq:hyper-encoder}
    \boldsymbol{h} = \mathcal{D}_r ({\rm ReLU}(\boldsymbol{W}_0 {\rm mean} (\boldsymbol{H}_{query}) + \boldsymbol{b}_0))
\end{equation}
where $\mathcal{D}_r$ denotes dropout.

The \textbf{decoder} of HyperExpert varies based on the structure of the target PEFT adapters.

\paragraph{Decoder of \method{}$_{LoRA}$} The decoder uses linear layers to transform the compressed representation $\boldsymbol{h}$ into the LoRA matrix for self-attention. For instance, the computations of LoRA matrix for $\boldsymbol{W}_q$ and $\boldsymbol{W}_k$ in self-attention are as follows:
\begin{equation}
\hat{\boldsymbol{A}}_q = \boldsymbol{W}_1\boldsymbol{h} + \boldsymbol{b}_1,
\hat{\boldsymbol{A}}_k = \boldsymbol{W}_2\boldsymbol{h} + \boldsymbol{b}_2.
\end{equation}
We only generate the $\boldsymbol{A}$ matrix in the LoRA module, initialize $\boldsymbol{B}$ to zero, and update it during training as the original LoRA setup. This ensures that $\Delta\boldsymbol{W} = \boldsymbol{B\hat{A}}$ starts from zero at the beginning of training.

\paragraph{Decoder of \method{}$_{Prompt}$} Prompt tuning includes an additional prompt embedding $\boldsymbol{E} \in \mathbb{R}^{K \times d}$ in each attention layer. The \textbf{decoder} is a linear layer that generates the prompt embedding $\boldsymbol{E}$ from the compressed query representation $\boldsymbol{h}$ as $\hat{\boldsymbol{E}} = \boldsymbol{W}_{p}\boldsymbol{h} + \boldsymbol{b}_p$, where $\boldsymbol{W}_{p} \in \mathbb{R}^{(K \times d) \times b}$. Here, $K$ is the prompt embedding length, and $d$ is the dimension of the transformer's hidden states. 

\paragraph{Decoder of \method{}$_{PAdapter}$} The parallel adapter is a bottleneck network composed of two linear layers. Therefore, our \textbf{decoder} uses two linear layers to generate the weights for the parallel adapter as follows:
\begin{equation}
\boldsymbol{L}_1 = \boldsymbol{W}_{l1}\boldsymbol{h} + \boldsymbol{b}_{l1},
\boldsymbol{L}_2 = \boldsymbol{W}_{l2}\boldsymbol{h} + \boldsymbol{b}_{l2}.
\end{equation}

\paragraph{Training Graph and Gradient Propagation} The entire \method{} framework is designed to be end-to-end differentiable. In the forward pass, the query is first processed by the initial $l$ transformer layers to obtain the hidden representation $\boldsymbol{H}_{query}^l$. This representation is fed into the Hypernetwork's encoder and decoder to dynamically generate the adapter parameters (e.g., $\hat{\boldsymbol{A}}_q, \hat{\boldsymbol{A}}_k$) for the subsequent $N-l$ layers. During the backward pass, the loss is computed based on the generated summary. The gradients flow back from the loss function through the output layers, directly into the generated adapter parameters in the final $N-l$ transformer layers. By the standard chain rule, the gradients are then propagated through the Hypernetwork's decoder and encoder, updating the Hypernetwork's own weights. Finally, the gradients continue to flow back through $\boldsymbol{H}_{query}^l$ to update the standard PEFT adapters in the initial $l$ layers. This continuous computational graph ensures that the hypernetwork is effectively optimized alongside the standard PEFT parameters in an end-to-end differentiable manner.


\paragraph{Implementation Details of Parameter Generation}
To clarify the generation process, the adapter parameters can be produced via two distinct computational graphs: \textbf{parallel generation} and \textbf{sequential generation}. 
\begin{itemize}
    \item \textbf{Parallel Generation}: In this approach, the Hypernetwork acts as a one-shot generator. It takes the query representation from the $l$-th layer ($\boldsymbol{H}_{query}^l$) and simultaneously outputs the complete set of adapter parameters required for all subsequent $N-l$ layers. This method allows the Hypernetwork's computation to be fully decoupled from the LLM's subsequent forward passes, highly optimizing generation latency.
    \item \textbf{Sequential Generation}: In contrast, this approach interleaves the parameter generation with the LLM's forward pass iteratively. Specifically, the query representation from the current $l$-th layer ($\boldsymbol{H}_{query}^l$) is used by the Hypernetwork to generate the adapter parameters for the very next layer, $(l+1)$. Once the LLM completes the forward pass for layer $(l+1)$, the newly updated $\boldsymbol{H}_{query}^{l+1}$ is fed back into the Hypernetwork to generate parameters for layer $(l+2)$, and so on. This layer-by-layer sequential conditioning allows the generated adapters to dynamically adjust to the evolving hidden states of the query.
\end{itemize}

In HyperExpert, the number of encoder layers can either match the number of target layers for which parameters are generated, or a single shared layer can be used. For the decoder, we adopt a shared-layer approach to reduce parameter overhead.

\subsection{Query-focused Infini-attention Module}\label{sec:qf-infini}

The Query-focused Infini-Attention mechanism consists of several key steps. The first step is \textbf{Fixed-length Local Attention}, designed to maximize the utilization of the capabilities of the self-attention mechanism. The input tokens are segmented for long context documents to perform standard causal dot-product attention within each segment. In both training and inference, we cache the previous segment's key-value (KV) attention states to pad the local attention KV states to a fixed length. The next step is \textbf{Compression and Retrieval of Memory}. Before completing the local attention for the current segment, the cached KV attention states are compressed into two memory blocks: one preserving the entire historical context and another retaining query-related information. These compressed memories are subsequently used to retrieve relevant context for the following segments. The final step is \textbf{Repeated Query Instruction}, which ensures accurate query-focused summarization during inference. To achieve this, we prepend and append the query instruction to the document. The prepended query instruction facilitates the compression of historical memory, while the appended query instruction ensures that local attention adheres to the full query instruction. 

\subsubsection{Memory compression}
For the $s$-th segment with length $L$, before computing the local attention,  we update the full context memory $\boldsymbol{M}_{s-1}^{all} \in \mathbb{R}^{d_{key} \times d_{value}}$ and the query-focused memory $\boldsymbol{M}_{s-1}^{query} \in \mathbb{R}^{d_{key} \times d_{value}}$, and a normalization term $\boldsymbol{z}_{s-1}\in \mathbb{R}^{d_{key}}$ is then used for memory retrieval as follows:
\begin{equation} \label{eq:Mall}
    \boldsymbol{M}_{s-1}^{all} \leftarrow \boldsymbol{M}_{s-2}^{all} + \sigma(\boldsymbol{K}_{cache})^T \boldsymbol{V}_{cache}
\end{equation}
\begin{equation} \label{eq:Mquery}
    \boldsymbol{M}_{s-1}^{query} \leftarrow \boldsymbol{M}_{s-2}^{query} + \sigma(\boldsymbol{K}_{cache})^T \hat{\boldsymbol{V}}_{cache}
\end{equation}
\begin{equation}
    \boldsymbol{z}_{s-1} \leftarrow \boldsymbol{z}_{s-2} + \sum_{t=1}^L \sigma(\boldsymbol{K}^t_{cache})
\end{equation}
where $\sigma$ is a nonlinear activation function. Following the work of \citet{katharopoulos2020transformersarernn} and \citet{munkhdalai2024infinitransformer}, we employ element-wise ELU+1 as the activation function \citep{clevert2015elus}. The term $\sigma(\boldsymbol{K})^T \boldsymbol{V}$ on the right side of Equation~\ref{eq:Mall} and~\ref{eq:Mquery} is referred to as an associative binding operator \citep{schlag2020associative}. The query-focused memory $\boldsymbol{M}_{s-1}^{query}$ differs from the full context memory only in the value states $\hat{\boldsymbol{V}}_{cache}$ used within the associative binding operator. We utilize the query states $\boldsymbol{Q}_{query}$ of query instruction to scale the value states and keep only query-related information $\hat{\boldsymbol{V}}_{cache}$ as
\begin{equation}
    \alpha_{i} = sigmoid \left(\frac{mean(\boldsymbol{Q}_{query})(\boldsymbol{K}_{cache}^i)^T}{\sqrt{d_{model}}} \right) 
\end{equation}
\begin{equation}
    \hat{\boldsymbol{V}}_{cache} = \boldsymbol{\alpha} \odot \boldsymbol{V}_{cache}.
\end{equation}
Here, we use the mean pooling of $\boldsymbol{Q}_{query}$ and the key states to compute a related score for each representation.

\subsubsection{Memory retrieval}
After updating the memory, we retrieve new content $\boldsymbol{\mathcal{A}}_{all} \in \mathbb{R}^{L \times d_{value}}$ and $\boldsymbol{\mathcal{A}}_{query} \in \mathbb{R}^{L \times d_{value}}$ from the full context memory $\boldsymbol{M}_{s-1}^{all}$ and the query-focused memory $\boldsymbol{M}_{s-1}^{query}$, respectively. This retrieval is performed using the query states $\boldsymbol{Q} \in \mathbb{R}^{L \times d_{key}}$ as follows:
\begin{equation}
    \boldsymbol{\mathcal{A}}_{all} = \frac{\sigma(\boldsymbol{Q})\boldsymbol{M}_{s-1}^{all}}{\sigma(\boldsymbol{Q})\boldsymbol{z}_{s-1}},     
    \boldsymbol{\mathcal{A}}_{query} = \frac{\sigma(\boldsymbol{Q})\boldsymbol{M}_{s-1}^{query}}{\sigma(\boldsymbol{Q})\boldsymbol{z}_{s-1}}
\end{equation}
\subsubsection{Long-term context injection}
First, we apply a linear layer to aggregate $\boldsymbol{\mathcal{A}}_{all}$ and $\boldsymbol{\mathcal{A}}_{query}$. Then, we aggregate the retrieved content and the local attention $\boldsymbol{\mathcal{A}}_{local}$ using a learned gating scalar $\boldsymbol{\beta}$:
\begin{equation}
    \boldsymbol{\gamma} = sigmoid(\boldsymbol{W}_g \boldsymbol{\mathcal{A}}_{query})
\end{equation}
\begin{equation}
    \boldsymbol{\mathcal{A}}_{ret} = \boldsymbol{\gamma} \odot \boldsymbol{\mathcal{A}}_{query} + (1-\boldsymbol{\gamma}) \odot \boldsymbol{\mathcal{A}}_{all}
\end{equation}
\begin{equation}
    \boldsymbol{\mathcal{A}} = sigmoid(\boldsymbol{\beta}) \odot \boldsymbol{\mathcal{A}}_{ret} + 
    (1-sigmoid(\boldsymbol{\beta})) \odot \boldsymbol{\mathcal{A}}_{local}
\end{equation}
where $\boldsymbol{W}_g \in \mathbb{R}^{1 \times d_{value}}$ is a trainable weight that dynamically merges the two retrieved contents. $\boldsymbol{\beta}$ contains a single scalar value per head as a training parameter, enabling a learnable trade-off between the long-term and local information flows in the model.

\subsubsection{Repeated query instruction}
To incorporate query instructions into the model, we concatenate the query instruction with the document as the model input. During local attention, the query states $\boldsymbol{Q}_{query}$ of the query instruction are utilized to compute query-focused memory within each segment. However, when generating summaries, the retrieved information from memory fails to effectively guide the model in producing summaries that adhere to the query instructions. To address this issue, we employ a straightforward approach: we replicate the query instruction at the end of the document. This ensures that the query instruction is within the window of the local attention computation when generating summaries, enabling the model to generate query-relevant summaries accurately.

\paragraph{The necessity of the Repeated query instruction and Potential Risks} While the repeated query instruction significantly improves QFS performance, a potential concern is whether it causes the model to overemphasize explicit the query and overlook indirectly relevant but crucial contextual information in the document. We argue that this strategy is practically essential, and its risks are structurally mitigated by our framework. Specifically, the initial query instruction is utilized within each local block using query states $\boldsymbol{Q}_{query}$ to compute the query-focused memory ($\boldsymbol{M}^{query}$) and perform memory retrieval from both the full context memory ($\boldsymbol{M}^{all}$) and $\boldsymbol{M}^{query}$. Once the retrieved content $\boldsymbol{\mathcal{A}}_{ret}$ and local attention $\boldsymbol{\mathcal{A}}_{local}$ are obtained and aggregated in the current block, the explicit constraint of the original query question tends to diminish or become "lost" in the subsequent forward pass. Therefore, the repeated query instruction acts as a crucial anchor for the QFS task. 

Furthermore, the risk of overlooking indirect information is safeguarded by our dual-memory architecture. Alongside the query-focused memory ($\boldsymbol{M}_{query}$), IDEAL explicitly retains the full context memory ($\boldsymbol{M}_{all}$), which captures the entire historical context without any query-biased filtering. During memory retrieval, the model accesses both $\boldsymbol{M}_{all}$ and $\boldsymbol{M}_{query}$ and aggregates them utilizing a learned gating scalar $\beta$. This structural design ensures that while the repeated query instruction guides the local attention towards the user's explicit intent, the globally important and indirectly relevant information preserved in the full context memory is effectively integrated and not overlooked.

\section{Experiments} \label{sec:experiments}

\subsection{Datasets}

We evaluate our approach on three query-focused summarization datasets: CovidET \citep{zhan2022CovidET}, QMSum \citep{zhong_qmsum_2021}, SQuALITY \citep{wang_squality_2022}. 
Table~\ref{table:datasets} shows the detailed statistics of the datasets used in our experiments. QMSum is a multi-domain dataset for meeting summarization, covering Product, Academic, and Committee meetings. QMSum(Golden) is a shorter version of QMSum where documents only contain sections relevant to the queries. SQuALITY is a public-domain dataset for story summarization. Unlike others, SQuALITY includes multiple summaries for each question. The input documents in the CovidET and QMSum (Golden) datasets have token counts of \textbf{228} and \textbf{2670}, respectively, when tokenized using the LLaMA2 tokenizer. In contrast, the QMSum and SQuALITY datasets feature longer input token lengths, with \textbf{8071} and \textbf{13227} tokens, respectively. 

\begingroup
\setlength{\tabcolsep}{3pt}
\begin{table*}[h]
  \centering
  \begin{tabular}{cccccccc}
    \toprule[1.5pt]
    \textbf{Type} & \textbf{Dataset} & \textbf{Domain} & \textbf{Instances} & \textbf{Input Tk}  & \textbf{Output Tk} & \textbf{Queries} & \textbf{Split} \\
    \hline
    \multirow{2}{*}{Query} & QMSum & Meeting & 1808 & 13227(2670$^*$) & 88 & 1566 & 1257/272/281 \\
    & SQuALITY(v1.3) & Story & 625 & 8071 & 306 & 437 & 1000/500/1040\\
    \multirow{1}{*}{Aspect}  & CovidET & Reddit & 7122 & 228 & 32 & 7 & 4188/1029/1524 \\
    \bottomrule[1.5pt]
  \end{tabular}
  \caption{\label{table:datasets}
    Statistics of query/aspect-based summarization datasets. \textbf{Instances} represents the total number of (document, summary) pairs in the corresponding dataset. \textbf{Input Tk} and \textbf{Output Tk} denote the number of input and output token lengths under the LLaMA2 tokenizer, respectively. \textbf{Queries} indicate the number of unique queries or aspects appearing in the dataset, respectively. 2670$^*$ represents the number of input tokens for QMSum(Golden). \textbf{Split} denotes the train/validation/test splits.
  }
\end{table*}
\endgroup

\subsection{Evaluation Metrics}
\subsubsection{Reference-based Evaluation Metric}
We evaluate the summaries using ROUGE metrics \citep{lin2004rouge}, including ROUGE-1, ROUGE-2, the sentence-level ROUGE-L, and the summary-level ROUGE-Lsum. Additionally, we use a Roberta-large version of BERTScore \citep{bertscore_2020}, which leverages Roberta-large to compute the similarity between the references and the model's outputs. Specifically, since SQuALITY includes multiple summaries for each question, we report multi-reference scores for all metrics following \citet{wang_squality_2022}. We calculate the metrics for each pair of a generated summary and multiple references, then choose the maximum score.


\subsubsection{LLM-based Evaluators}
Recent studies, such as G-Eval \cite{liu2023geval} and GPTRank \citep{liu2024onlearning}, demonstrate that LLM-based evaluators outperform reference-based metrics with higher alignment to human judgments. Using GPTRank, we evaluate our method by prompting the LLM to first generate an \textbf{explanation} and then provide a \textbf{ranking} for a list of candidate summaries corresponding to the same source article.

\subsection{Implementation Details}
We use the pre-trained LLaMA (2-7B, 3.1-8B) \citep{touvron2023llama} with \( N = 32 \) transformer layers as the backbone model. LLaMA3.1-8B served as our primary baseline model. However, for low-memory experiments with Query-focused Infini-attention, the more memory-efficient LLaMA2-7B was employed.
All \method{} models are trained by the AdamW optimizer with a cosine annealing schedule after the warmup starts. The warmup epochs, batch size, learning rate, and weight decay are set to 1, 32, 0.006, and 0.02, respectively. We use the validation set to find the optimal epochs for each dataset. During the generation stage, we adopt top-p sampling as the default decoding method with a temperature of 0.1 and a top-p value of 0.75. 

For \method{}$_{Prompt}$, we follow LLaMA-Adapter-v1 \citep{zhang_llama-adapter_2023}, adopting a prompt length \( K = 10 \) and applying prompts to the last 30 layers, with the prompts of the last 15 layers are generated. For \method{}$_{PAdapter}$, adapters are applied to the first 16 layers and generated for the last 16 layers. For \method{}$_{LoRA}$, only the \( \boldsymbol{A} \) matrix in the LoRA module is generated for the last 16 layers. 

All LLaMA-based models in our experiments use Automatic Mixed Precision, with 16-bit for frozen parameters and 32-bit for trainable parameters to conserve memory. Additionally, we employ Flash-Attention2 \citep{dao2023flashattention2} to accelerate model training and inference for LLaMA-based models. All models in our experiments can be trained on at least a single 24GB Nvidia GeForce RTX 3090, except for the large local context size setting for long documents. 


For the BART model baselines, we use the HuggingFace Transformers library and the AdamW optimizer. We set the batch size, learning rate, and weight decay to 32, 0.0001, and 0.1, respectively.


\subsection{Comparison of Methods}


Our approach is compared against several fully fine-tuned pre-trained language models frequently employed for summarization, including BART-large \citep{lewis_bart_2019}, LED-base-OASum \citep{yang_oasum_2023}, and HMNet \citep{zhu2020HMNet} (with results reported by \citet{zhong_qmsum_2021}). We further evaluate the LLaMA (3.1 8B) and three corresponding PEFT-based baselines: Prompt, PAdapter, and LoRA. Comparisons are also conducted with GPT-4O (version of 2024-08-06). 

For long document datasets, we compare our methods against several Retrieval-Augmented Generation (RAG) approaches, including BART-large + DPR \citep{wang_squality_2022}, HMNet + Locator \citep{zhong_qmsum_2021}, and Qontsum \citep{Sotudeh2023QontSumOC}. We also include the abstractive summarizer SegEnc \citep{vig2022exploringNeuQF}, its pre-training framework Socratic Pretraining \citep{pagnoni2023socratic}, and Unlimiformer \citep{bertsch2024unlimiformer}, a retrieval-based method for handling unlimited-length inputs.




\begingroup
\setlength{\tabcolsep}{3pt}
\begin{table*}[t]
  \centering
  \begin{tabular}{lcccccccccccc}
    \toprule[1.5pt]
    & \multicolumn{6}{c}{\texttt{CovidET  Dataset}} & \multicolumn{6}{c}{\texttt{QMSum(Golden)  Dataset}} \\
    \hline
    \hline
    \textbf{Models}  & \textbf{LC} & \textbf{R-1} & \textbf{R-2} &  \textbf{R-L} & \textbf{R-Lsum} & \textbf{BScore} & \textbf{LC} & \textbf{R-1} & \textbf{R-2} &  \textbf{R-L} & \textbf{R-Lsum} & \textbf{BScore}\\
    
    \hline
    Bart-large & 1K & 27.54 & 7.72 & 21.66 & 22.24 & 88.61 & 1K & 38.49 & 14.26 & 25.25 & 33.75 & 86.38\\
    LED-base-OASum$^*$ & 4K & 25.61 & 6.58  &- & 20.45 & - &-&-&-&-&-&-\\
    HMNet$^*$ &-&-&-&-&-&- & - & 36.06 & 11.36 & - & 31.27 & - \\
    ChatGPT$^*$ & - & 20.81 & 3.99 & 15.35 & 15.36 & - & - & 36.83 & 12.78 & 24.23 & 24.19 & -\\
    GPT-4O & 1K & 17.01 & 3.35 & 13.18 & 13.24 & 87.33 & 3K & 33.46 & 10.22 & 20.59 & 30.18 & 85.87\\
    Llama3.1-8B & 1K &12.85 & 2.12 & 9.32 & 11.04 & 84.55 & 3K &20.51 & 6.76 & 13.94 & 18.44 & 82.39\\
    Prompt & 1K & 29.18 & 8.77 & \colorbox{second}{\textbf{23.64}} & \colorbox{second}{\textbf{24.15}} & \colorbox{second}{\textbf{89.20}} & 3K & 34.81 & 13.33 & 24.99 & 30.73 & 86.69 \\ 
    PAdapter & 1K & 29.37 & \colorbox{second}{\textbf{8.84}} & 23.20 & 23.86 & 89.06 & 3K & 39.41 & 15.77 & 28.18 & 34.98 & 87.54 \\ 
    Lora & 1K & 29.00 & 8.43 & 22.79 & 23.43 & 89.00 & 3K & 40.69 & 16.11 & 28.84 & 36.18 & 87.71 \\ 
    \hline
    \method{}$_{Prompt}$ & 1K & 29.10 & \colorbox{best}{\textbf{9.01}} & \colorbox{best}{\textbf{23.65}} & \colorbox{best}{\textbf{24.18}} & \colorbox{best}{\textbf{89.26}} & 3K & 35.58 & 13.75 & 25.33 & 31.52  & 86.73\\ 
    \method{}$_{PAdapter}$ & 1K & \colorbox{second}{\textbf{29.51}} & 8.78 & 23.21 & 23.80 & 89.07 & 3K & \colorbox{second}{\textbf{40.79}} & \colorbox{best}{\textbf{17.24}} & \colorbox{second}{\textbf{29.64}} & \colorbox{second}{\textbf{36.55}}  & \colorbox{second}{\textbf{87.80}} \\ 
    \method{}$_{LoRA}$ & 1K & \colorbox{best}{\textbf{29.62}} & \colorbox{second}{\textbf{8.84}} & 23.40 & 24.06 & 89.12 & 3K & \colorbox{best}{\textbf{40.85}} & \colorbox{second}{\textbf{16.89}} & \colorbox{best}{\textbf{29.67}} & \colorbox{best}{\textbf{36.66}} & \colorbox{best}{\textbf{87.83}}\\ 
    \bottomrule[1.5pt]
  \end{tabular}
  \caption{\label{table:CovidET and QMSum(Golden)}
    Comparison with baselines on CovidET and QMSum(Golden). \textbf{LC} denotes the local context size of the model. \textbf{R-L}, \textbf{R-Lsum}, and \textbf{BScore} denote ROUGE-L, ROUGE-Lsum, BERTSCore, respectively. $^*$ indicates that experimental results are obtained from related work. We color each row as the \colorbox{best}{\textbf{best}} and \colorbox{second}{\textbf{second best}}. 
  }
\end{table*} %
\endgroup

\subsection{Main Results of \method{}}

\begingroup
\setlength{\tabcolsep}{3pt}
\begin{table*}[t]
  \centering
  \begin{tabular}{lcccccc cccccc}
  \toprule[1.5pt]
    & \multicolumn{6}{c}{\texttt{SQuALITY Dataset}} & \multicolumn{6}{c}{\texttt{QMSum Dataset} } \\
    \hline
    \hline
    \textbf{Models} & \textbf{LC} & \textbf{R-1} & \textbf{R-2} &  \textbf{R-L} & \textbf{R-Lsum} & \textbf{BScore} & \textbf{LC} & \textbf{R-1} & \textbf{R-2} &  \textbf{R-L} & \textbf{R-Lsum} & \textbf{BScore}\\
    
    \hline
    Bart-large & 1K & 38.58 & 9.81 & 20.97 & 36.11 & 84.81 & 1K  & 31.76 & 7.76 & 20.02 & 27.52 & 85.22 \\
    LED-base-OASum$^*$ & 4K & 37.6 & 8.81 & - & 35.14 & - & 4K & 30.30 & 7.56 & - & 26.67 & - \\
    Bart-Large+DPR$^*$ & 1K & 41.5 & 11.4 & 21.0 & - & 85.5 &-&-&-&-&-&- \\
    HMNet+Locator$^*$ &-&-&-&-&-&-  & - & 32.29 & 8.67 & - & 28.17 & - \\ 
    ChatGPT$^*$ & - &  37.02 & 8.19 & 18.45 & 22.56 & - & - & 28.34 & 8.74 & 17.81 & 18.81 & -  \\
    GPT-4O & 8K & 41.01 & 9.84 & 21.02 & 36.99 & 85.39 & 16K & 30.30 & 8.13 & 18.34 & 26.80 & 84.40\\
    Llama3.1-8B & 8K  & 36.84 & 9.34 & 20.14 & 34.17 & 84.39 & 16K & 21.34 & 6.26 & 14.39 & 19.03 & 82.69 \\
    Bart+Unlimiformer$^*$ &-&-&-&-&-&- & 1K & 30.9 & 8.0 & 19.9 & - & - \\
    SegEnc$^*$ & - & 45.68 & 14.51 & 22.47 & - & 85.86 & - & 37.05 & 13.03 & - & 32.62 & 87.44 \\
    SegEnc+Socratic Pret.$^*$ & - & 46.31 & 14.80 & 22.76 & - & 86.04 & - & 38.06 & 13.74 & - & 33.51 & 87.63 \\
    Qontsum$^*$ & - & 45.76 & 14.27 & 24.14 & - & 86.07 & - & \colorbox{second}{\textbf{38.42}} & 13.50 & - & \colorbox{second}{\textbf{34.03}} & 87.72 \\
    Prompt & 8K & 36.81 & 10.72 & 23.69 & 33.26 & 85.23 & 8K & 30.52 & 9.77 & 21.50 & 26.33 & 85.89\\ 
    PAdapter & 8K & 43.86 & 13.13 & 24.21 & 40.83 & 86.55 & 8K & 36.03 & 12.61 & 24.64 & 31.74 & 86.96 \\ 
    Lora & 8K & 44.47 & 13.23 & 24.32 & \colorbox{second}{\textbf{41.46}} & 86.63 & 8K & 36.40 & 12.10 & 23.98 & 31.87 & 86.66 \\ 
    \hline
    \method{}$_{Prompt}$ & 8K & 37.40 & 11.23 & 23.93 & 34.44 & 85.56 & 8K & 31.41 & 10.60 & 22.27 & 27.19 & 86.08 \\ 
    \method{}$_{PAdapter}$ & 8K & 44.37 & 13.43 & \colorbox{second}{\textbf{24.76}} & \colorbox{best}{\textbf{41.47}} & \colorbox{second}{\textbf{86.66}} & 8K & 37.27 & \colorbox{second}{\textbf{13.90}} & \colorbox{best}{\textbf{26.32}} & 32.73 & 87.19  \\ 
    \method{}$_{LoRA}$ & 8K & 43.87 & 13.86 & \colorbox{best}{\textbf{25.54}} & 40.99 & \colorbox{best}{\textbf{86.86}}  & 8K & \colorbox{best}{\textbf{38.67}} & \colorbox{best}{\textbf{14.42}} & \colorbox{second}{\textbf{26.28}} & \colorbox{best}{\textbf{34.24}} & 87.29  \\ 
  \bottomrule[1.5pt]
  \end{tabular}
  \caption{\label{table:QMSum and SQuALITY}
    Comparison with baselines on two long document datasets, SQuALITY and QMSum. \textbf{LC} denotes the local context size of the model. \textbf{R-L}, \textbf{R-Lsum}, and \textbf{BScore} denote ROUGE-L, ROUGE-Lsum, BERTSCore, respectively. $^*$ indicates that experimental results are obtained from related work. We color each row as the \colorbox{best}{\textbf{best}} and \colorbox{second}{\textbf{second best}}
  }
\end{table*} 
\endgroup

Tables~\ref{table:CovidET and QMSum(Golden)}-~\ref{table:QMSum and SQuALITY} present the results on QFS datasets. Our approaches achieve the best results overall. \method{} consistently outperforms the corresponding PEFT Adapters. For instance, on the QMSum(Golden) dataset, \method{}$_{PAdapter}$ surpasses PAdapter by 1.46 (5.2$\%$) ROUGE-L points and 1.57 (4.5$\%$) ROUGE-Lsum points with the same input size of 3K.

For the two long document datasets shown in Table~\ref{table:QMSum and SQuALITY}, \method{}$_{LoRA}$ with an input length of 8K achieved the best ROUGE-L and BERTScore on SQuALITY, and the best ROUGE-1, ROUGE2, and ROUGE-Lsum on QMSum dataset. \method{}$_{PAdater}$ achieved the best ROUGE-Lsum on SQuALITY, and the best ROUGE-L on QMSum dataset.





\subsection{LLM-based Evaluation}
\begingroup
\begin{table}[h]
\small
  \centering
  \begin{tabular}{lcccccc}
  \toprule[1.5pt]
    \textbf{Models} & \textbf{Win} & \textbf{Lose} & \textbf{Tie} &  \textbf{R-L} & \textbf{R-Lsum} & \textbf{BScore} \\
    \hline
    \hline
    \multicolumn{7}{c}{\texttt{SQuALITY Dataset}} \\
    \hline
    Socratic Pret & 39 & 214 & 1 & 23.14 & \textbf{42.28} & 85.86 \\ 
    \hline
    \method{}$_{LoRA}$ & \textbf{214} &  \textbf{39} & 1 & \textbf{25.54} & 40.99 & \textbf{86.86} \\ 
  \bottomrule[1.5pt]
  \end{tabular}
  \caption{
    GPTRank comparison between Socaric Pret. and \method{}$_{LoRA}$.
  }
  \label{table:GPTRank}
\end{table} 
\endgroup

The results on reference-based metrics indicate that our method achieves a certain level of effectiveness. However, while these metrics are simple and fast, they suffer from poor correlation with human evaluators, lack interpretability, and fail to capture high-level semantic qualities of summaries. To address this, we employ GPTRank (GPT-4O) to compare our method with the open-source state-of-the-art (SOTA) approach, Socratic Pret \citep{pagnoni2023socratic}, for evaluating the high-level semantic qualities of summaries.

As shown in the table~\ref{table:GPTRank}, while Socratic Pret achieves comparable performance to \method{}$_{LoRA}$ on ROUGE-L and BERTScore and even surpasses \method{}$_{LoRA}$ on ROUGE-Lsum, the comparison using GPTRank reveals a different perspective. Across 254 test samples, \method{}$_{LoRA}$ outperforms Socratic Pret in 214 cases, achieving an 84\%  win rate. This indicates that although the SOTA non-LLM method performs similarly to LLM-based methods on certain metrics, it lags significantly in terms of high-level semantic qualities. Furthermore, analysis of the GPTRank evaluations reveals that summaries generated by Socratic Pret exhibit more inconsistencies with the original article compared to those generated by \method{}$_{LoRA}$. In contrast, \method{}$_{LoRA}$ produces more concise and fluent summaries.

Table~\ref{tab:case_study} presents a comprehensive example comparing \method{}$_{LoRA}$ and Socratic Pret using GPTRank, including the full prompt and GPT's evaluation response, except the full article due to space consideration. We evaluate the quality of the two summaries via prompt engineering with GPT, providing an explanation, a one-word reason for inferior summaries, and an indication of the superior summary (or a tie). To mitigate potential bias from summary ordering, the order of the two summaries was randomized in our experiments.

\begin{table*}[ht]
    \centering
    \begin{tabular}{p{1\linewidth}}
        \toprule[1.5pt]
         \textbf{Prompt:}   \\
        \hline
        You will be provided with an article along with a query instruction and two summaries that respond to the query instruction, numbered as follows: 1. Summary 1, 2. Summary 2. Please compare the quality of these two summaries and determine which one is better (a tie is also possible). First, provide an explanation for your decision. Then, specify a one-word reason for why the less favorable summary is inferior. Finally, give your decision in the format: 1, 2, or tie. Response format: \textbf{Explanation}: “Your explanation here”.  \textbf{Reason}: “One word reason”.  \textbf{Decision}: 1 or 2 or tie. \\
        \textbf{Query Instruction}: {What is Gurn's role in the story?} \\ \textbf{Article}: Produced by Greg Weeks, Mary Meehan and the Online Distributed Proofreading Team at http://www.pgdp.net Raiders of the Second Moon By GENE ELLERMAN A strange destiny had erased Noork's memory,and had brought him to this tiny world to write ... \\ 
        \textbf{Summary 1}: Gurn is the leader of the wild Vasads of the jungle. He is a renegade from the walled city of Grath, and the city rulers call him a traitor because he has spoken against the making of slaves of the captured Zurans from other valleys. \colorbox{red}{Gurn is also an exile from the city of Konto}, and he has told \colorbox{red}{Noork that his brother}, Gurn, has told him the reason for the exile. The reason is that the Zurans are no longer making slaves of other valleys, and Gurn believes that they should no longer make them slaves. \colorbox{red}{He tells Noork to go to Gurn} and tell him that the Misty Ones can be trapped and skinned. \\
        \textbf{Summary 2}: Gurn is the leader of the Vasads, a group of people who live in the jungles of Sekk. He is a tall, golden-skinned man with a bracelet of golden discs linked together with human hair. He is an exile from the walled city of Grath, and is considered a traitor by the city rulers. He is Noork’s friend, and Noork tells Sarna that he has been friends with Gurn. Gurn is also the one who gave Noork his short powerful bow.\\
        \hline
        \hline
        \textbf{GPT Response:} \\ 
        \hline
        \textbf{Explanation}: Summary 2 provides a more accurate and concise depiction of Gurn's role in the story. It highlights his leadership position among the Vasads, his physical description, his exile status from Grath, his friendship with Noork, and his contribution to Noork by gifting him a bow. On the other hand, Summary 1 contains factual inaccuracies, confuses Gurn's relationship with other characters, and includes irrelevant information, such as exiles and slavery, that is not central to understanding Gurn's role. \textbf{Reason}: “Inaccuracy”. \textbf{Decision}: 2. \\
        \bottomrule[1.5pt]
    \end{tabular}
    \caption{An example comparing summaries generated by \method{}$_{LoRA}$ and Socratic Pret using GPTRank on SQuALITY dataset. The red-highlighted text indicates significant errors. In this example, Summary 1 is generated by Socratic Pret, and Summary 2 is generated by \method{}$_{LoRA}$.}
    \label{tab:case_study}
\end{table*}

\subsection{Performance of Low Memory \method{}} 

\begingroup
\begin{table*}[t]
  \centering
  \begin{tabular}{lcccc cccc}
     \toprule[1.5pt]
    \textbf{Models} & \multicolumn{4}{c}{\texttt{QMSum Dataset}} & \multicolumn{4}{c}{\texttt{SQuALITY Dataset}} \\
    \hline
    \hline
    & LC  & R-L & R-Lsum & BScore & LC & R-L & R-Lsum & BScore \\
    \hline
    Lora & 1.6K & 19.58 & 25.25 & 84.93 & 1.6K &  20.73 & 34.41 & 85.31 \\ 
    \hline
    \multirow[t]{3}{*}{\method{}$_{LoRA}$}  & 1.6K & 19.71 & 26.27 & 85.29 & 1.6K & 21.16 & 34.73 & 85.52\\  
     & 3.8K & 21.62 & 28.46 & 85.94 & 3.8K &  22.54 & 37.54 & 85.83\\
     & 8K &  26.28 & 34.24 & 87.29 & 8K & 25.54 & 40.99 & 86.86 \\
     \hline
    LoRA+Inf & 0.8/6K & 21.13 & 26.58 & 86.00 & 1.6/9K & 20.59 & 34.76 & 85.02 \\
    \method{}$_{LoRA}$+Inf & 0.8/6K  & 21.76 & 26.16 & 86.02 & 1.6/9K &  21.68 & 34.81 & 85.28 \\ 
    \makecell[r]{w/o ReQ} & 0.8/6K  & 16.57 & 20.40 & 84.37 & 1.6/9K &  17.89 & 30.62 & 84.13 \\
    \method{}$_{LoRA}^{QF\_Inf}$ & 0.8/6K  & 22.16 & 27.05 & 86.16 & 1.6/9K & 21.49 & 34.86 & 85.54 \\  
   \bottomrule[1.5pt]
  \end{tabular}
  \caption{\label{table:ablationQM_SQ}
    Comparing \method{}$_{LoRA}^{QF\_Inf}$ with Infini-attention based methods and \method{}$_{LoRA}$ with different input size. LoRA+Inf and \method{}$_{LoRA}$+Inf denote the incorporation of Infini-attention into LoRA and \method{}$_{LoRA}$, respectively. w/o ReQ indicates that the query instruction is not repeated at the end of the input document. 
  }
\end{table*} 
\endgroup

\method{}$_{LoRA}$ consistently demonstrates improved performance as training input length increases. However, this comes at the cost of increased GPU memory consumption. Table~\ref{table:ablationQM_SQ} illustrates this trade-off, showcasing \method{}$_{LoRA}$ performance on input lengths of 1.6K, 3.8K, and 8K, requiring 24G, 40G, and 80G of memory, respectively. In contrast to \method{}$_{LoRA}$, our proposed \method{}$_{LoRA}^{QF\_Inf}$ that integrated with Query-focused Infini-attention exhibits memory efficiency when handling long inputs. \method{}$_{LoRA}^{QF\_Inf}$ maintains a consistent memory footprint of 24G regardless of the input length. Notably, on the QMSum dataset, \method{}$_{LoRA}^{QF\_Inf}$ outperforms \method{}$_{LoRA}$ with an input length of 1.6K on all metrics within the same 24GB memory constraint. Moreover, it surpasses \method{}$_{LoRA}$ with an input length of 3.8K in 40GB memory on the ROUGE-L metric. 

\subsection{Ablation Study}

\subsubsection{\method{}$_{LoRA}$ vs LoRA by different training sequence length}



\begin{figure}[t]
    \centering
    \includegraphics[width=0.5\textwidth, trim=0.05in 0.05in 0.05in 0.05in, clip]{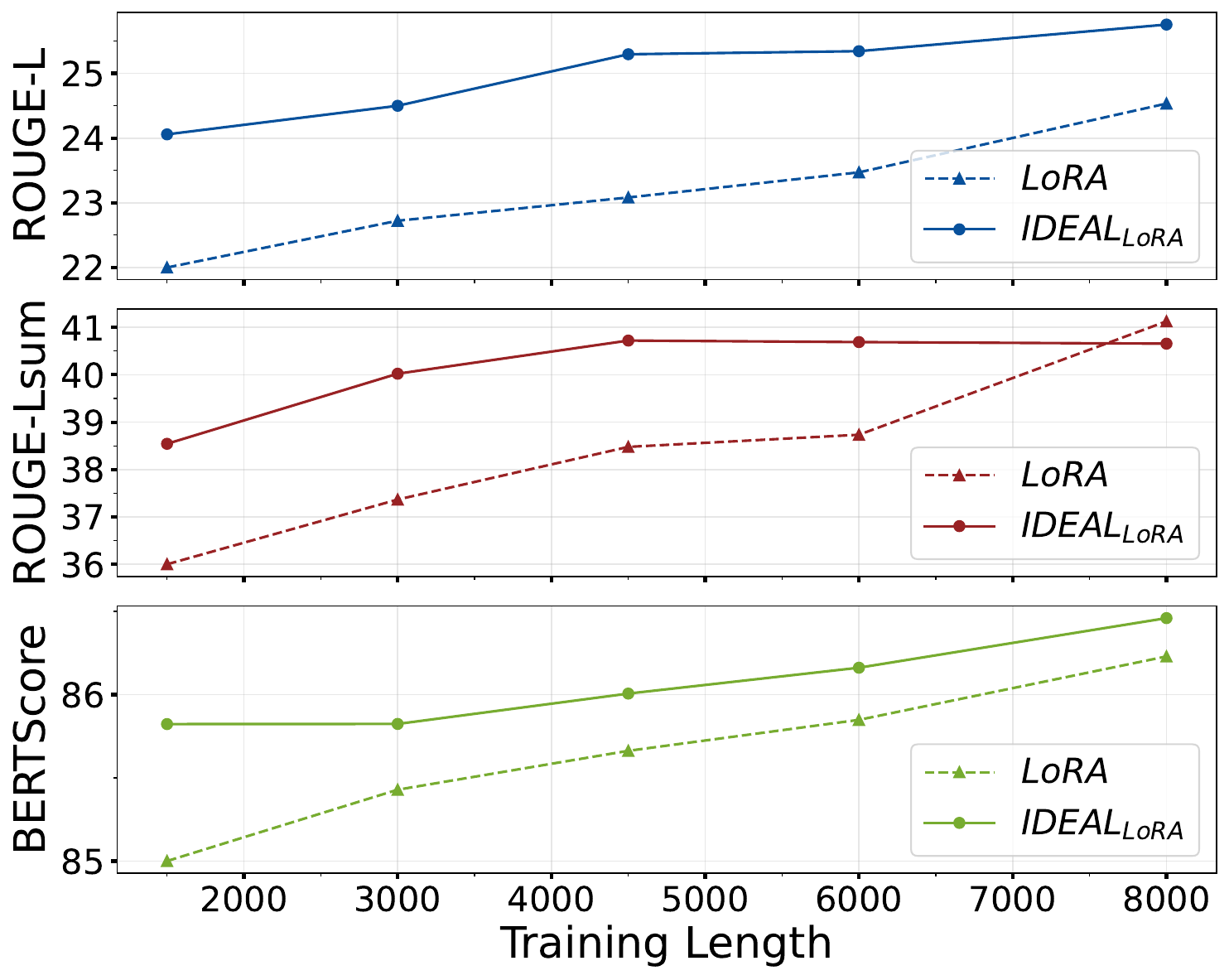}
    \caption{A comparison of LoRA and \method{}$_{LoRA}$ under different training sequence lengths on SQuALITY dataset.
}
    \label{fig:lora_Ideal_seqlen}
    \vspace{-3mm}
\end{figure}

To evaluate the effectiveness of our approach under varying training sequence lengths, we compared \method{}$_{LoRA}$ and LoRA on the SQuALITY dataset across training lengths ranging from 1500 to 8000. Figure~\ref{fig:lora_Ideal_seqlen} illustrates the results in terms of ROUGE-L, ROUGE-Lsum, and BERTScore metrics. The results demonstrate that our method consistently improves performance on the QFS task across different sequence lengths.

\subsubsection{Comparative Analysis of HyperExpert Variants: Performance and Parameter Efficiency}

\begin{figure}[htbp]
    \centering
    \includegraphics[width=0.5\textwidth, trim=0.05in 0.05in 0.05in 0.05in, clip]{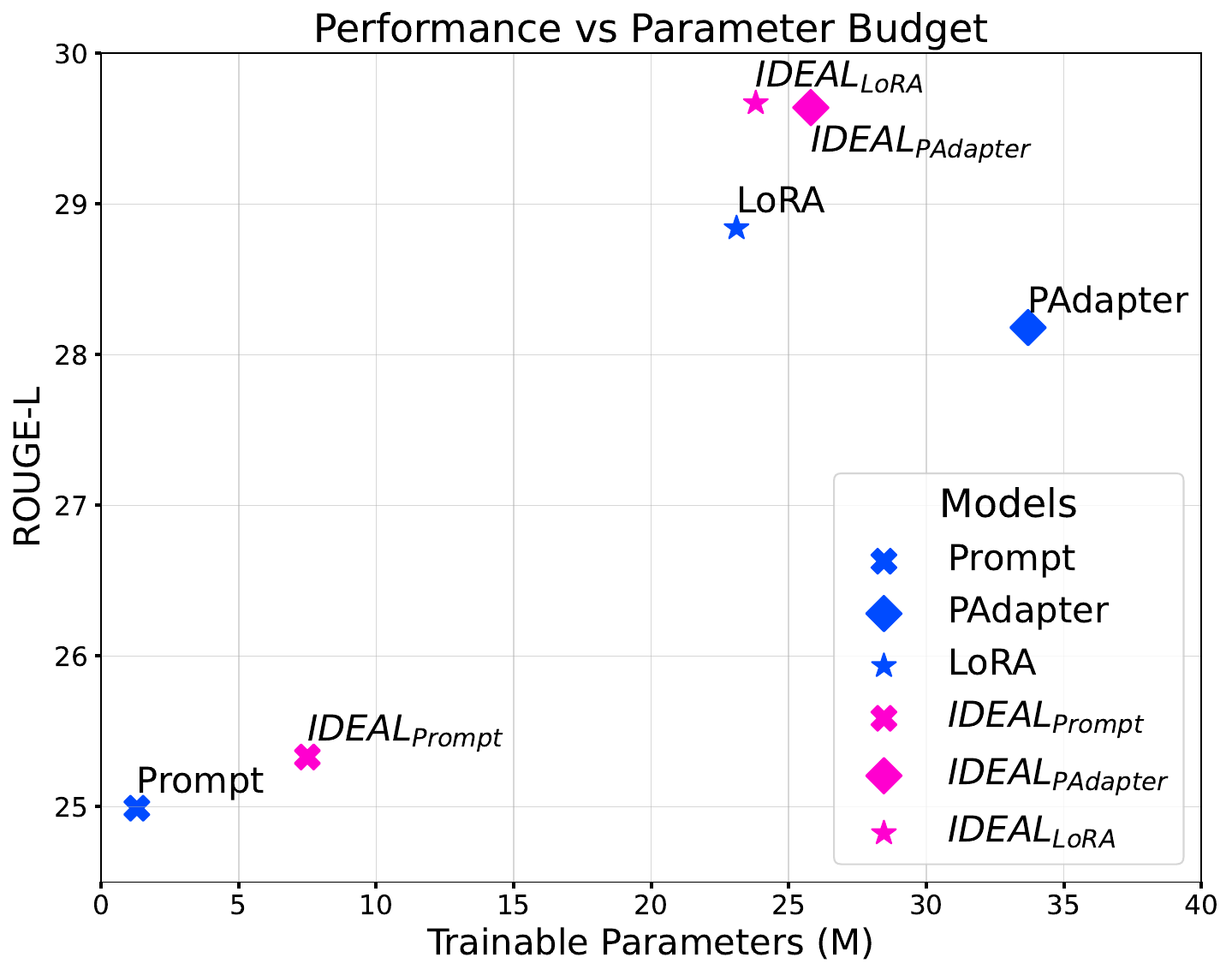}
    \caption{Our Query-aware HyperExperts outperform the corresponding PEFT methods on QFS tasks using a comparable amount of trainable parameters on QMSum(golden) dataset.
}
    \label{fig:performance_parameters}
    \vspace{-3mm}
\end{figure}

\begingroup
\begin{table*}[htbp]
  \centering
  \begin{tabular}{lcccccc}
    \toprule[1.5pt]
    \textbf{Models} & \textbf{Prompt} & \textbf{PAdapter} & \textbf{LoRA} & \textbf{\method{}$_{Prompt}$} & \textbf{\method{}$_{PAdapter}$} & \textbf{\method{}$_{LoRA}$} \\
    \midrule
    \textbf{Params (M)} & 1.31 & 33.69 & 23.07 & 7.52 & 25.76 & 23.76 \\
    \bottomrule[1.5pt]
  \end{tabular}
  \caption{Trainable parameter sizes of different models.}
  \label{table:model_parameters}
\end{table*}
\endgroup


Table~\ref{table:model_parameters} outlines the trainable parameter counts for the three baseline PEFT models and their corresponding HyperExpert variants. Notably, compared to the standard baselines, the HyperExpert variants, specifically \method{}$_{PAdapter}$ and \method{}$_{LoRA}$ either maintain or even reduce the overall number of trainable parameters. Concurrently, as illustrated in Figure~\ref{fig:performance_parameters}, all three HyperExpert variants demonstrate improved performance over their respective baselines, with \method{}$_{PAdapter}$ and \method{}$_{LoRA}$ exhibiting particularly significant gains. This indicates that our proposed Query-aware HyperExpert method effectively enhances multiple baseline PEFT approaches without introducing excessive computational overhead. 

When comparing the three variants against each other, their relative performance aligns with the intrinsic capabilities of their corresponding baselines. \method{}$_{Prompt}$ requires the fewest parameters but also yields the lowest performance among the three. This is primarily because Prompt-tuning adjusts the parameter count solely by controlling the prompt length, and merely increasing the prompt length does not guarantee proportional performance improvements \citep{hu_llm-adapters_2023}. Between the remaining two robust variants, \method{}$_{LoRA}$ achieves superior summarization performance compared to \method{}$_{PAdapter}$ while utilizing fewer parameters. Combining these observations with the main results presented in Tables~\ref{table:CovidET and QMSum(Golden)}  and \ref{table:QMSum and SQuALITY}, \method{}$_{LoRA}$ emerges as the best-performing variant overall. 
\paragraph{Model Selection Guidelines} For end users, we recommend \method{}$_{LoRA}$ as the primary choice for most QFS scenarios due to its optimal balance of high performance and parameter efficiency. \method{}$_{PAdapter}$ serves as a strong alternative when specific architectural constraints favor feed-forward network adaptations. Finally, \method{}$_{Prompt}$ is suitable only for scenarios operating under extreme parameter or memory restrictions where a slight degradation in summary quality is acceptable.


\subsubsection{The layers to generate parameters}
Table~\ref{table:layers_generate} presents the results of \method{}$_{LoRA}$ on the QMSum (Golden) dataset when generating LoRA parameters for different numbers of layers using a hypernetwork. The results indicate that generating parameters for the last 16 layers achieves the best performance.

\begingroup
\setlength{\tabcolsep}{2pt}
\begin{table}[h]
\small
  \centering
  \begin{tabular}{ccccccc}
    \toprule[1.5pt]
    \textbf{Layers} & \textbf{R-1} & \textbf{R-2} & \textbf{R-L} & \textbf{R-Lsum} & \textbf{BScore} & \textbf{Params(M)} \\
    \hline
    8-32 & 40.04 & 16.56 & 29.06 & 35.71 & 87.77 & 24.55 \\
    16-32 & \colorbox{best}{\textbf{40.85}} & \colorbox{best}{\textbf{16.89}} & \colorbox{best}{\textbf{29.67}} & \colorbox{best}{\textbf{36.66}} & \colorbox{best}{\textbf{87.83}} & 23.76 \\ 
    24-32 & 40.73 & 16.36 & 28.96 & 36.22 & 87.76 & 22.98 \\
    - & 40.69 & 16.11 & 28.84 & 36.18 & 87.71 & 23.07 \\
    \bottomrule[1.5pt]
  \end{tabular}
  \caption{\label{table:layers_generate}
    Different number of layers that the LoRA parameters are generated of \method{}$_{LoRA}$ on \textbf{QMSum(Golden)} dataset. \textbf{16-32} indicates that the LoRA parameters from layers 16 to 32 are generated by the Hypernetwork. \textbf{-} indicates no generated parameters.
  }
\end{table}
\endgroup

\subsubsection{Different configuration of HyperExpert} 

Table~\ref{table:hyperexpert_config} shows the experimental results for four adapter parameter generation configurations: parallel versus sequential generation, and using different or shared encoders within HyperExpert. Parallel generation with different encoders yields the best performance.

\begingroup
\setlength{\tabcolsep}{2pt}
\begin{table}[h]
\small
  \centering
  \begin{tabular}{cccccccc}
    \toprule[1.5pt]
    \textbf{Gen} & \textbf{Enc} & \textbf{R-1} & \textbf{R-2} & \textbf{R-L} & \textbf{R-Lsum} & \textbf{BScore} & \textbf{Params(M)} \\
    \hline
    Para & Diff & \colorbox{best}{\textbf{40.85}} & \colorbox{best}{\textbf{16.89}} & \colorbox{best}{\textbf{29.67}} & \colorbox{best}{\textbf{36.66}} & \colorbox{best}{\textbf{87.83}} & 23.76 \\ 
    Seq & Diff & 40.49 & 16.56 & 28.95 & 35.99 & 87.70 & 23.76 \\
    Seq & Share & 40.57 & 16.40 & 29.01 & 36.25 & 87.76 & 19.83 \\
    Para & Share & 40.43 & 16.62 & 28.82 & 36.18 & 87.76 & 19.83\\
    \bottomrule[1.5pt]
  \end{tabular}
  \caption{\label{table:hyperexpert_config}
  Four configurations for generating adapter parameters on \textbf{QMSum (Golden)} are evaluated: \textbf{Para} denotes parallel generation, while \textbf{Seq} refers to sequential generation. \textbf{Diff} indicates that the encoder in HyperExpert corresponds to the number of transformer layers generating parameters, whereas \textbf{Share} represents the use of a single shared encoder layer.
  }
    
\end{table}
\endgroup

\subsubsection{The diversity of generated parameters}
\begin{figure}[ht]
\begin{subfigure}[h]{0.5\textwidth}
\includegraphics[width=1\linewidth]{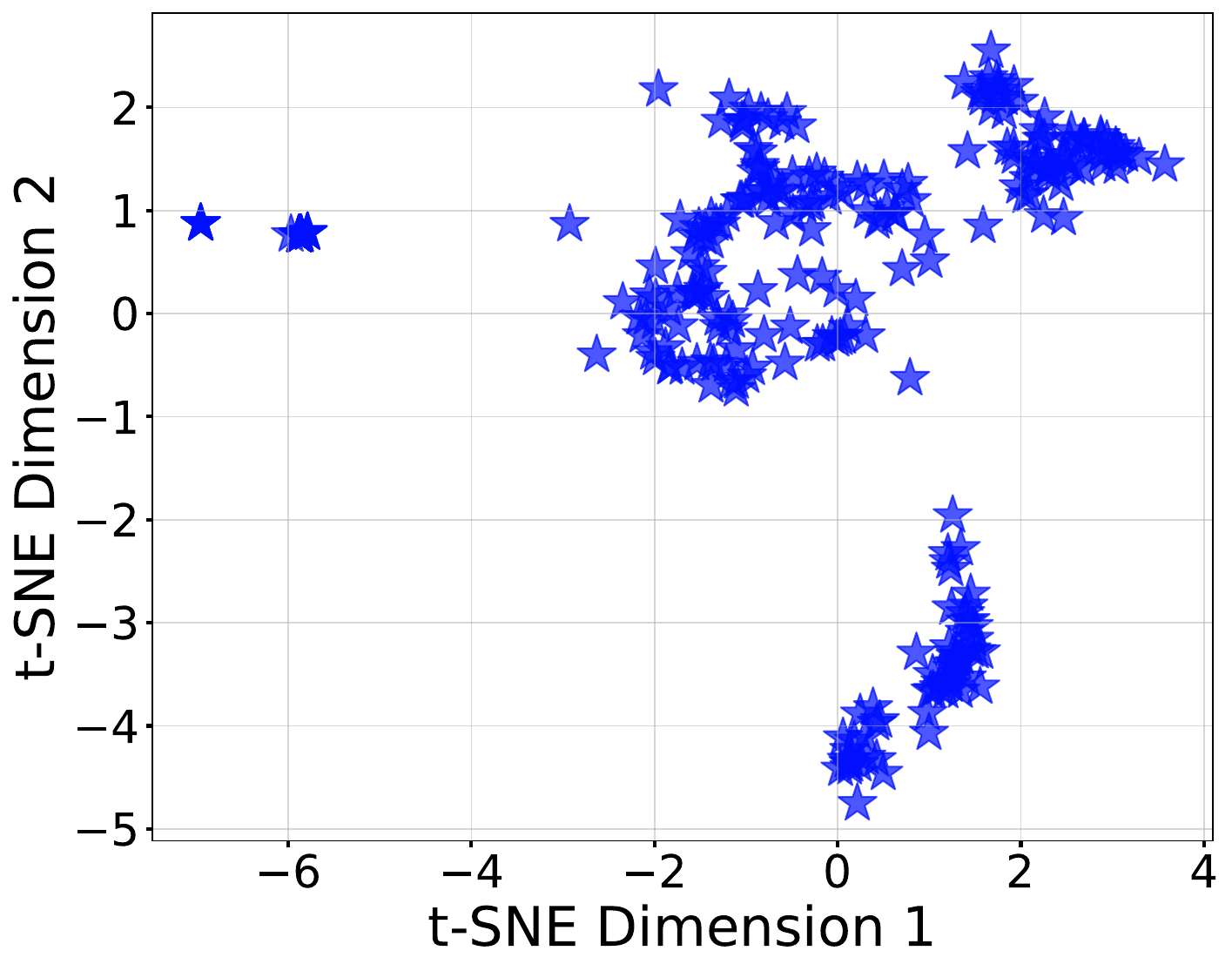} 
\caption{QMSum.}
\label{fig:gen_params_tsne_QMSum}
\end{subfigure}
\hspace{1mm}
\begin{subfigure}[h]{0.5\textwidth}
\includegraphics[width=1\linewidth]{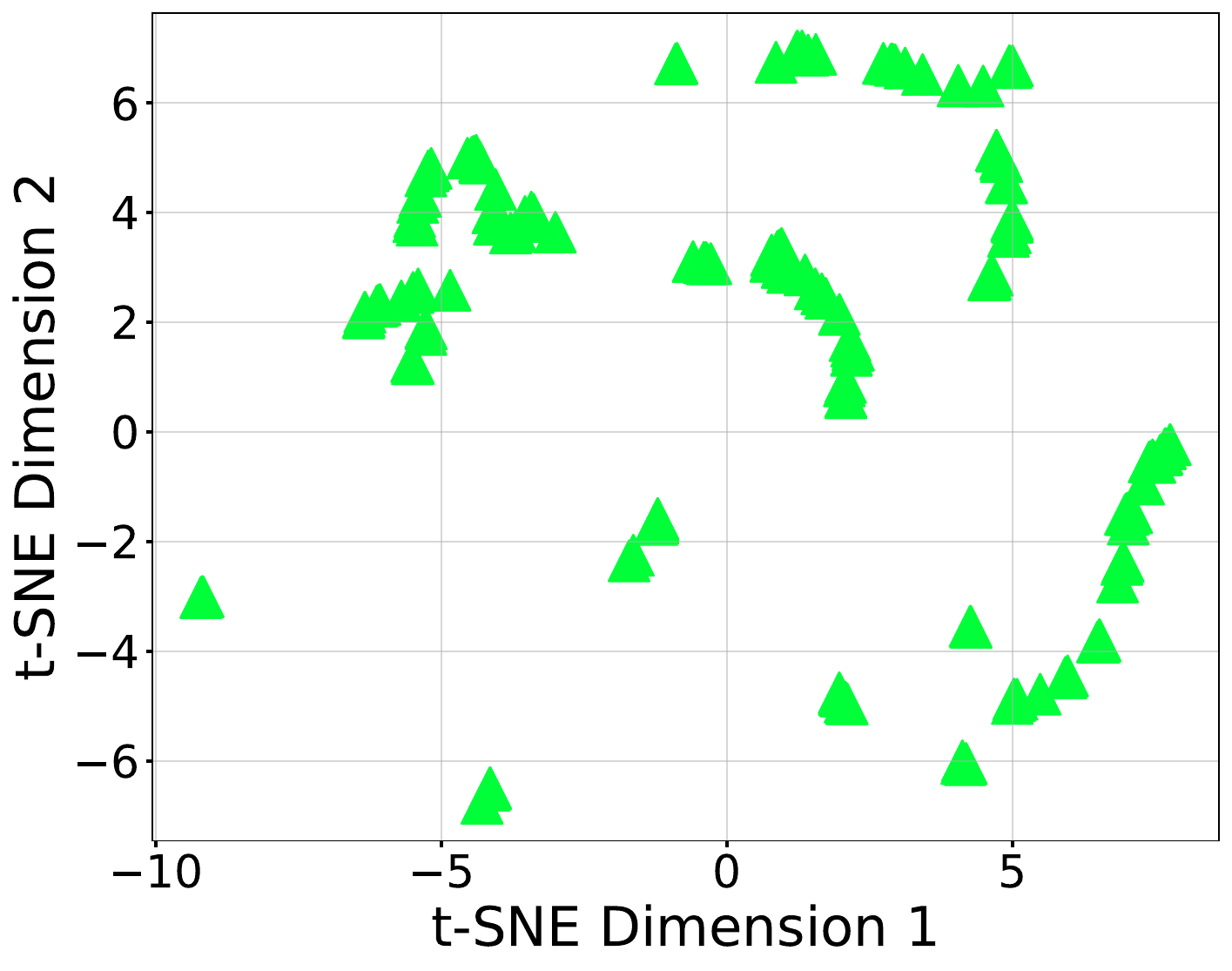}
\caption{SQuALITY.}
\label{fig:gen_params_tsne_SQuALITY}
\end{subfigure}

\caption{t-SNE Visualization of Query-based Parameters' Dynamic Characterizations.}
\label{fig:gen_params_tsne}
\end{figure}

To intuitively illustrate the diversity of parameters generated by the \method{} model, we applied the t-SNE algorithm to visualize the adapter parameters of a selected layer in two dimensions on the QMSum and SQuALITY test sets. As shown in Figure~\ref{fig:gen_params_tsne}, the generated parameters exhibit distinct distributions across the two datasets, with clearly identifiable clusters. This demonstrates that IDEAL can dynamically generate adapter parameters conditioned on the query.

\subsubsection{The input representation of HyperExpert} 
Our HyperExpert takes the representation of the query instruction as input. For comparison, we conducted experiments using the representations of the document and the combined query and document as inputs. The results in table~\ref{table:hyper_input} show that using only the query representation as input achieves the best performance, while using the combined query and document representation yields the second-best results.

\begingroup
\setlength{\tabcolsep}{3pt}
\begin{table}[t]
\small
  \centering
  \begin{tabular}{cccccc}
    \toprule[1.5pt]
    \textbf{Input} & \textbf{R-1} & \textbf{R-2} & \textbf{R-L} & \textbf{R-Lsum} & \textbf{BScore} \\
    \hline
    Query & \colorbox{best}{\textbf{40.85}} & \colorbox{best}{\textbf{16.89}} & \colorbox{best}{\textbf{29.67}} & \colorbox{best}{\textbf{36.66}} & \colorbox{best}{\textbf{87.83}}\\ 
    Document & 40.17 & 16.10 & 28.84 & 35.84 & 87.70 \\
    Query\&Document & 40.23 & 16.47 & 29.14 & 35.84 & 87.60 \\
    \bottomrule[1.5pt]
  \end{tabular}
  \caption{\label{table:hyper_input}
  Different input representation of \method{}$_{LoRA}$ on \textbf{QMSum(Golden)}.
  }
    
\end{table}
\endgroup

\subsubsection{The effectiveness of each module in \method{}$_{LoRA}^{QF\_Inf}$}

In Table~\ref{table:ablationQM_SQ}, we evaluated the effectiveness of Query-focused Infini-attention through comparative testing. First, we implemented Infini-attention based on LoRA as Lora+Inf and observed significant improvements compared to LoRA alone under the same GPU memory constraints, with increases of 1.55 and 1.33 points in ROUGE-L and ROUGE-Lsum on the QMSum dataset, respectively. These results indicate that compressing the key-value states of historical segments enables the summarization of long documents within limited GPU memory. Furthermore, we enhanced \method{}$_{LoRA}$ with Infini-attention, achieving better results than Lora+Inf in ROUGE-L. The \method{}$_{LoRA}^{QF\_Inf}$ outperformed both \method{}$_{LoRA}$+Inf and Lora+Inf in all metrics, demonstrating that our proposed Query-focused Infini-attention effectively compresses query-related information. For the \method{}$_{LoRA}$+Inf method, we observed a significant decline in all metrics after removing the repeated query instruction at the end of the input document, demonstrating the necessity of repeating the query instruction.

\subsubsection{Local context size of \method{}$_{LoRA}^{QF\_Inf}$}
Figure~\ref{fig:diff_localcontext} presents the performance of \method{}$_{LoRA}^{QF\_Inf}$ under varying local context sizes (LC). On the QMSum dataset, the model exhibits stable performance when LC exceeds 400, achieving nearly the best overall performance at LC=800. Similarly, on the SQuALITY dataset, the optimal LC is observed at 1.6K. These findings indicate that \method{}$_{LoRA}^{QF\_Inf}$ differs from \method{}$_{LoRA}$, the limited memory for the former is enough to handle extremely long inputs.

\subsubsection{Max input length of \method{}$_{LoRA}^{QF\_Inf}$}
\begin{figure}[t]
\begin{subfigure}[h]{0.5\textwidth}
\includegraphics[width=1\linewidth]{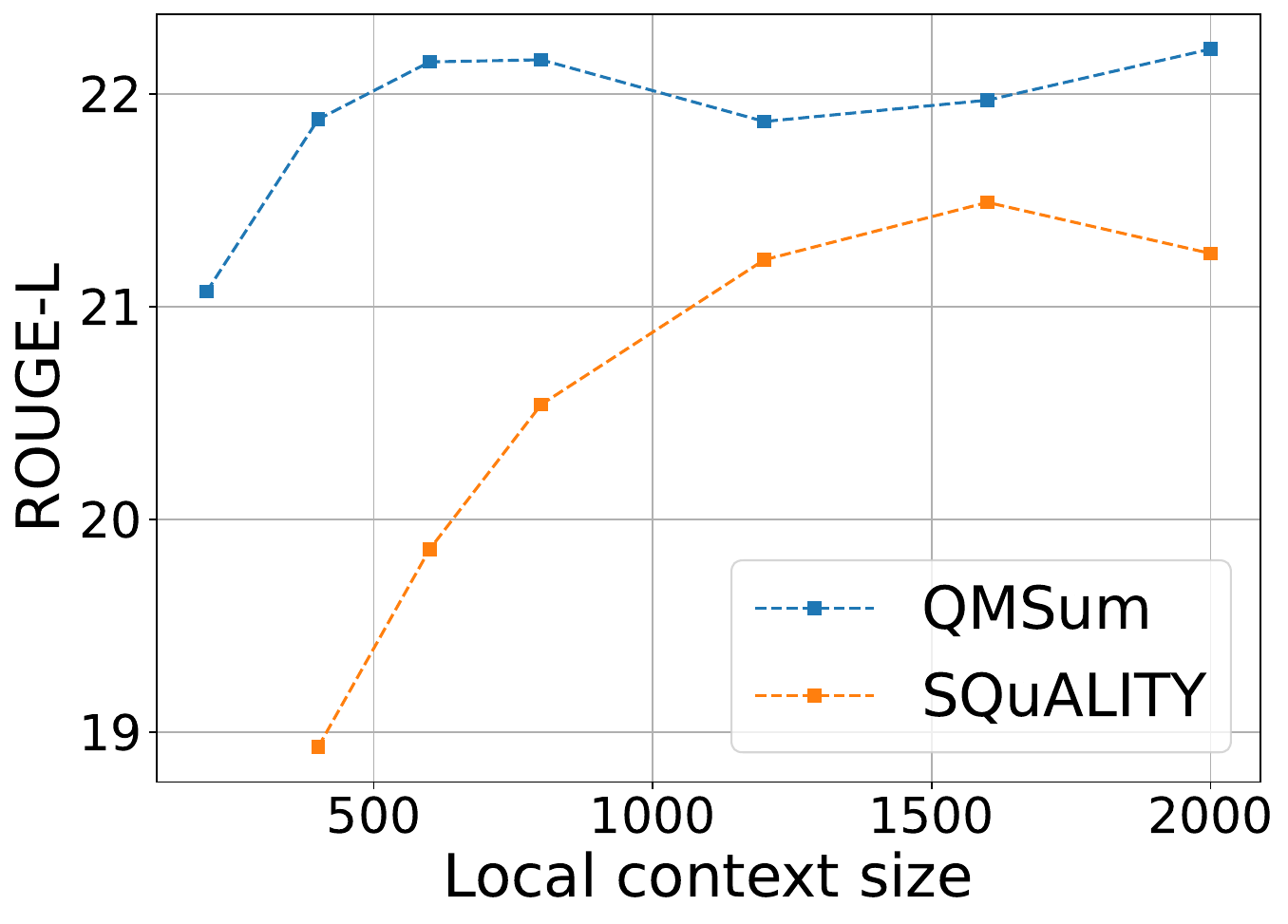} 
\caption{Performance with respect to the different local context size of \method{}$_{LoRA}^{QF\_Inf}$.}
\label{fig:diff_localcontext}
\end{subfigure}
\hspace{1mm}
\begin{subfigure}[h]{0.5\textwidth}
\includegraphics[width=1\linewidth]{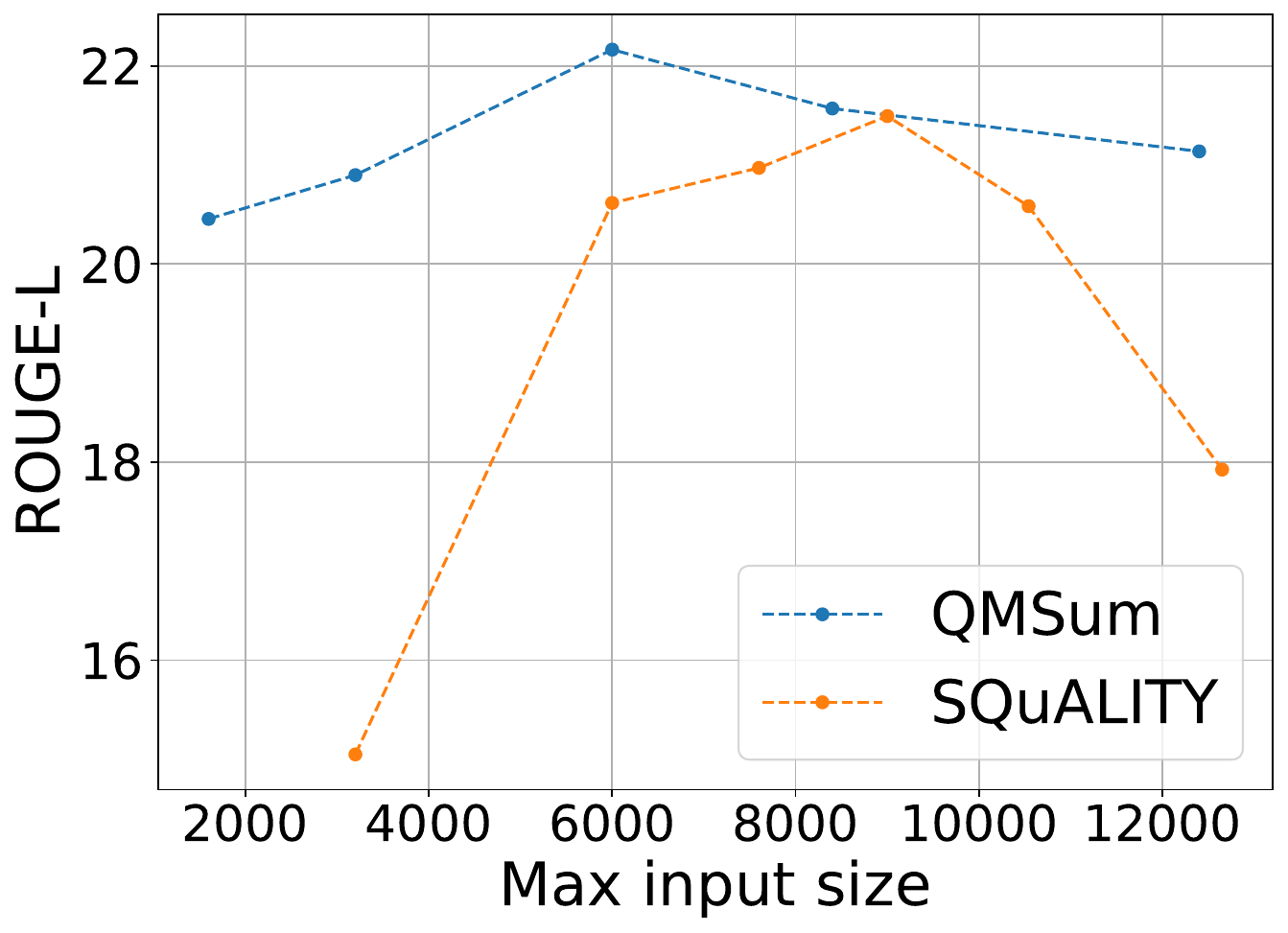}
\caption{Performance with respect to the different max input length of \method{}$_{LoRA}^{QF\_Inf}$.}
\label{fig:diff_max_length}
\end{subfigure}

\caption{Local and max input length of \method{}$_{LoRA}^{QF\_Inf}$.}
\label{fig:two_picture}
\end{figure}

Figure~\ref{fig:diff_max_length} presents the optimal max input length for \method{}$_{LoRA}^{QF\_Inf}$ on the QMSum and SQuALITY datasets. The results suggest that information relevant to the query in the QMSum dataset is primarily concentrated within the first 6000 tokens, while in the SQuALITY dataset, the relevant information is more evenly distributed throughout the document.



\subsection{The performance under LLaMA2 and LLaMA3 backbones.}

Table~\ref{table:backbone_comparison} shows the comparison of \method{}$_{LoRA}$ under LLaMA2-7B and LLaMA3-8B, LLaMA3.1-8B backbones with the same 24GB GPU memory. Due to the higher memory consumption of \method{}$_{LoRA}$ with the LLaMA 3 series backbone, resulting in a smaller local context size, evaluations show that it performs better with the LLaMA2-7B backbone given the same GPU memory. Therefore, in the low-memory experiments presented in Table~\ref{table:ablationQM_SQ}, LLaMA2-7B was used for all models except \method{}$_{LoRA}$ with an 8K context length, which used LLaMA 3.1-8B.

\begingroup
\begin{table*}[t]
  \centering
  \begin{tabular}{lccccccccc}
    \toprule[1.5pt]
    \textbf{Models} & Backbone & \textbf{LC} & \textbf{R-1} & \textbf{R-2} & \textbf{R-L} & \textbf{R-Lsum} & \textbf{BScore} & \textbf{Params(M)} \\
    \hline
    \method{}$_{LoRA}$ & LLaMA2-7B & 1.6K & \textbf{40.82} & \textbf{16.61} & \textbf{29.00} &  \textbf{36.08} & \textbf{87.68} & 24.5 \\
    \method{}$_{LoRA}$ & LLaMA3-8B & 1K & 39.99 & 15.63 & 27.89 & 35.34 & 87.54 & 23.8 \\ 
    \method{}$_{LoRA}$ & LLaMA3.1-8B & 1K & 39.90 & 16.06 & 28.55 & 35.57 & 87.48 & 23.8 \\
    \bottomrule[1.5pt]
  \end{tabular}
  \caption{\label{table:backbone_comparison}
    The comparison under LLaMA2 and LLaMA3, LLaMA3.1 backbones on \textbf{QMSum(Golden)} dataset with 24GB GPU memory.
  }
\end{table*}
\endgroup

\subsection{Performance Evaluation using the Qwen3-8B Backbone.}

\begingroup
\setlength{\tabcolsep}{3pt}
\begin{table*}[t]
  \centering
  \begin{tabular}{lcccccc cccccc}
  \toprule[1.5pt]
    & \multicolumn{6}{c}{\texttt{SQuALITY Dataset}} & \multicolumn{6}{c}{\texttt{QMSum Dataset} } \\
    \hline
    \hline
    \textbf{Models} & \textbf{LC} & \textbf{R-1} & \textbf{R-2} &  \textbf{R-L} & \textbf{R-Lsum} & \textbf{BScore} & \textbf{LC} & \textbf{R-1} & \textbf{R-2} &  \textbf{R-L} & \textbf{R-Lsum} & \textbf{BScore}\\
    \hline
    Prompt & 8K & 39.76 & 13.90 & 24.81 & 37.45 & 86.41 & 8K & 35.01 & 11.95 & 24.43 & 29.47 & 86.76\\ 
    PAdapter & 8K & 44.98 & 14.33 & 25.01 & 42.11 & 87.27 & 8K & 37.93 & 14.02 & 25.88 & 32.96 & 87.28 \\ 
    Lora & 8K & 45.03 & 14.81 & 25.29 & 42.86 & 87.47 & 8K & 38.08 & 14.01 & 25.90 & 33.58 & 87.46 \\
    \hline
    \method{}$_{Prompt}$ & 8K & 40.55 & 14.81 & 25.71 & 39.07 & 87.86 & 8K & 36.82 & 13.11 & 24.95 & 31.21 & 87.58 \\ 
    \method{}$_{PAdapter}$ & 8K & \colorbox{second}{\textbf{46.73}} & \colorbox{second}{\textbf{15.82}} & \colorbox{second}{\textbf{25.99}} & \colorbox{best}{\textbf{43.03}} & \colorbox{second}{\textbf{87.86}} & 8K & \colorbox{best}{\textbf{39.33}} & \colorbox{second}{\textbf{14.97}} & \colorbox{best}{\textbf{27.51}} & \colorbox{best}{\textbf{35.52}} & \colorbox{second}{\textbf{87.59}}  \\ 
    \method{}$_{LoRA}$ & 8K & \colorbox{best}{\textbf{46.88}} & \colorbox{best}{\textbf{15.89}} & \colorbox{best}{\textbf{26.78}} & \colorbox{second}{\textbf{42.90}} & \colorbox{best}{\textbf{88.21}}  & 8K & \colorbox{second}{\textbf{39.27}} & \colorbox{best}{\textbf{15.22}} & \colorbox{second}{\textbf{27.35}} & \colorbox{second}{\textbf{35.04}} & \colorbox{best}{\textbf{87.92}}  \\ 
  \bottomrule[1.5pt]
  \end{tabular}
  \caption{\label{table:QMSum and SQuALITY on Qwen3-8B}
    Performance comparison of the three HyperExpert variants and their corresponding PEFT baselines on the SQuALITY and QMSum datasets utilizing the Qwen3-8B backbone. \textbf{LC} denotes the local context size of the model. \textbf{R-L}, \textbf{R-Lsum}, and \textbf{BScore} denote ROUGE-L, ROUGE-Lsum, BERTSCore, respectively. We color each row as the \colorbox{best}{\textbf{best}} and \colorbox{second}{\textbf{second best}}.
  }
\end{table*}
\endgroup

To further verify the cross-model generalization ability of our proposed \method{} method, we conducted supplementary experiments utilizing another mainstream open-source LLM with a non-LLaMA architecture, Qwen3-8B \citep{yang2025qwen3}. The evaluations were performed on two long document datasets: SQuALITY and QMSum. As shown in Table~\ref{table:QMSum and SQuALITY on Qwen3-8B}, all three HyperExpert variants achieve significant performance improvements over their corresponding standard PEFT baselines on the Qwen3-8B backbone. These consistent gains effectively demonstrate that the efficacy of the IDEAL framework is not limited to the LLaMA family, but generalizes robustly across different LLM architectures.

\subsection{Training Time Comparison}

Table~\ref{table:training_time} presents a detailed comparison of training times between our proposed methods and the baselines. Notably, all three HyperExpert variants introduce no additional training time overhead when compared to their respective standard PEFT baselines. Among these variants, \method{}$_{Prompt}$ is the most time-efficient, requiring the least training time. Additionally, the integration of our Query-focused Infini-attention module in \method{}$_{LoRA}^{QF\_Inf}$ results in only a slight increase in training time compared to both \method{}$_{LoRA}$+Inf and LoRA+Inf.

\begin{table}[t]
\small
  \centering
  \begin{tabular}{lcccccc}
    \toprule[1.5pt]
    \textbf{Models} & \textbf{LC} & \textbf{Time/Epoch} &  \\
    \hline
    Prompt &  1.6K & 9min \\
    \method{}$_{Prompt}$ & 1.6K & 9min \\ 
    PAdapter &  1.6K & 12min \\
    \method{}$_{PAdapter}$ & 1.6K & 11min \\ 
    Lora &  1.6K & 11min \\
    \method{}$_{LoRA}$ & 1.6K & 11min \\ 
    \hline
    LoRA+Inf & 0.8/6K & 45min \\ 
    \method{}$_{LoRA}$+Inf & 0.8/6K & 46min \\
    \method{}$_{LoRA}^{QF\_Inf}$ & 0.8/6K & 50min  \\
    \bottomrule[1.5pt]
  \end{tabular}
  \caption{\label{table:training_time}
    Training time per epoch with 2 Nvidia GeForce RTX 3090 GPUs in data parallel mode on the QMSum dataset.
  }
\end{table} 

\subsection{Human Evaluation}

\begingroup
\setlength{\tabcolsep}{3pt}
\begin{table}[t]
\scriptsize
    \centering
    \begin{tabular}{lcccccc}
    \toprule[1.5pt]
        Models & \textbf{LC} & \textbf{Best} & \textbf{Worst} & \textbf{Correctness} & \textbf{Coverage} &  \textbf{Rouge-L} \\
        \hline
        Bart-Large & 1K & 33 & 75 & 3.3 & 2.9 & 25.25\\
        \hline
        LoRA(LLaMA2-7B)& 1.6K & 52 & 46 & 3.6 & 3.4 & 27.36\\
        \hline
        \method{}$_{LoRA}$(LLaMA2-7B) & 1.6K & 65 & 29 & 3.7 & 3.6 & 29.00 \\
    \bottomrule[1.5pt]
    \end{tabular}
    \caption{The human evaluation results of model-generated summaries obtained using the same 24G memory GPU on the QMSum (Golden) dataset. We summed the votes of the three evaluators for the Best and Worst metrics. For Correctness and Coverage, we first calculated the average for all samples, and then computed the mean across the three evaluators.}
    \label{tab:human-evaluation}
\end{table}
\endgroup

\begin{figure}[t]
    \centering
    \includegraphics[width=0.5\textwidth, trim=0.05in 0.05in 0.05in 0.05in, clip]{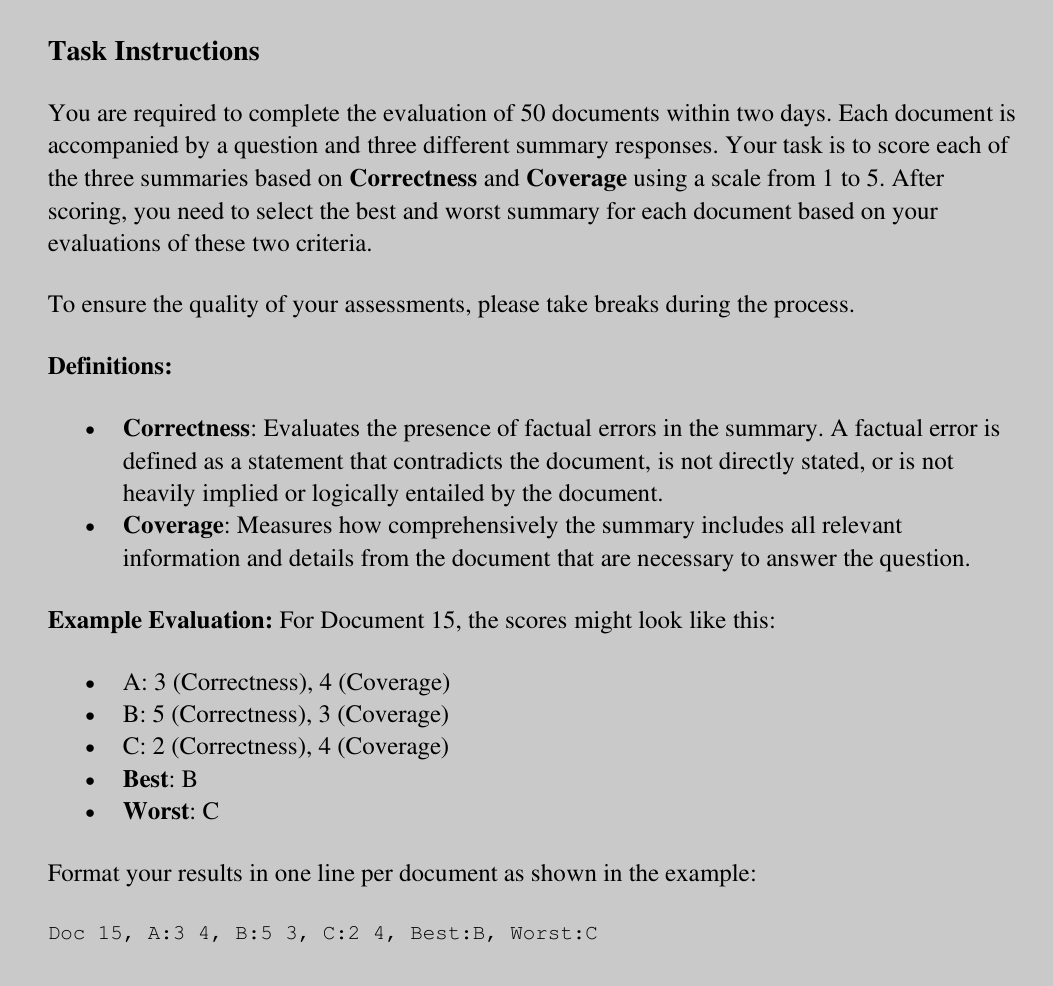}
    \caption{Task instructions of Human evaluation.}
    \label{fig:task-instruction}
\end{figure}

To ensure a fair evaluation, we conducted a human evaluation of the summaries generated by BART-Large, LoRA (LLaMA2-7B), and IDEAL$_{LoRA}$ (LLaMA2-7B) on the QMSum (Golden) dataset, all using identical computational resources. Each model was tested using a 3090 (24GB) GPU. We recruited three well-educated evaluators and randomly selected 50 samples to evaluate the summary quality of the three models from two aspects: Correctness and Coverage \citep{wang_squality_2022}. For each sample, the evaluators read the document and the corresponding question, then selected the best and worst summary among the three. In each case, we randomized the order of the summaries from the three models. The task instruction are detailed in Figure~\ref{fig:task-instruction}.

As shown in table~\ref{tab:human-evaluation}, although the BART-Large model did not lag far behind the other two models in terms of the ROUGE-L metric, it received significantly fewer "Best" summary votes and the most "Worst" votes. This may be because the LLaMA-based methods benefit from the understanding and reasoning capabilities of LLMs. IDEAL$_{LoRA}$ received 13 more "Best" votes compared to LoRA and had the fewest "Worst" votes, far less than LoRA's 46 and BART-Large's 75. This demonstrates that our proposed Query-focused PEFT method is indeed effective for QFS tasks.






\section{Conclusion}
In this paper, we propose \method{}, an efficient query-aware adaptation method on LLMs for QFS tasks, which consists of two modules: Query-aware HyperExpert and Query-focused Infini-attention. 
The two modules enable LLMs to achieve fine-grained query-LLM alignment efficiently and have the ability to handle lengthy documents. 
Experimental results demonstrate that our method improves performance on reference-based metrics. Furthermore, in pairwise comparisons against the SOTA fine-tuning method, Socratic Pret \citep{pagnoni2023socratic}, using the LLM-based evaluator GPTRank, our method achieved a win probability of 0.84, demonstrating its effectiveness.



\appendix
\section{List of Abbreviations} 
\label{app1}

\begin{table}[htbp]
\centering
\begin{tabular}{ll}
\toprule
\textbf{Abbreviation} & \textbf{Description} \\
\midrule
QFS & Query-Focused Summarization \\
LLM & Large Language Model \\
IDEAL & \textbf{I}nfinite and \textbf{D}ynamic larg\textbf{E} langu\textbf{A}ge mode\textbf{L} \\
PEFT & Parameter-Efficient Fine-Tuning \\
LoRA & Low-Rank Adaptation \\
PAdapter & Parallel Adapter \\
KV & Key-Value \\
LC & Local Context Size \\
R-1 / R-2 & ROUGE-1 / ROUGE-2 \\
R-L & ROUGE-L \\
R-Lsum & ROUGE-Lsum \\
BScore & BERTScore \\
ReQ & Repeated Query Instruction \\
Inf & Infini-attention \\
QF\_Inf & Query-focused Infini-attention \\
DPR & Dense Passage Retrieval \\
NLP & Natural Language Processing \\
\bottomrule
\end{tabular}
\caption{List of Abbreviations used in this paper.}
\label{tab:abbreviations}
\end{table}

\clearpage


 \bibliographystyle{elsarticle-num-names} 
 \bibliography{custom}

@inproceedings{liu2023geval,
  title={G-Eval: NLG Evaluation using Gpt-4 with Better Human Alignment},
  author={Liu, Yang and Iter, Dan and Xu, Yichong and Wang, Shuohang and Xu, Ruochen and Zhu, Chenguang},
  booktitle={Proceedings of the 2023 Conference on Empirical Methods in Natural Language Processing},
  pages={2511--2522},
  year={2023}
}

@inproceedings{liu2024onlearning,
  title={On Learning to Summarize with Large Language Models as References},
  author={Liu, Yixin and Shi, Kejian and He, Katherine and Ye, Longtian and Fabbri, Alexander Richard and Liu, Pengfei and Radev, Dragomir and Cohan, Arman},
  booktitle={Proceedings of the 2024 Conference of the North American Chapter of the Association for Computational Linguistics: Human Language Technologies (Volume 1: Long Papers)},
  pages={8639--8656},
  year={2024}
}

@inproceedings{lin2004rouge,
  title={Rouge: A package for automatic evaluation of summaries},
  author={Lin, Chin-Yew},
  booktitle={Text summarization branches out},
  pages={74--81},
  year={2004}
}

@inproceedings{dao2023flashattention2,
  title={Flash{A}ttention-2: Faster Attention with Better Parallelism and Work Partitioning},
  author={Dao, Tri},
  booktitle={International Conference on Learning Representations (ICLR)},
  year={2024}
}

@article{zhang2024hyperllava,
  title={HyperLLaVA: Dynamic Visual and Language Expert Tuning for Multimodal Large Language Models},
  author={Zhang, Wenqiao and Lin, Tianwei and Liu, Jiang and Shu, Fangxun and Li, Haoyuan and Zhang, Lei and Wanggui, He and Zhou, Hao and Lv, Zheqi and Jiang, Hao and others},
  journal={arXiv preprint arXiv:2403.13447},
  year={2024}
}

@article{daume2009bayesian,
  title={Bayesian query-focused summarization},
  author={Daum{\'e} III, Hal},
  journal={arXiv preprint arXiv:0907.1814},
  year={2009}
}

@article{gambhir2017recent,
  title={Recent automatic text summarization techniques: a survey},
  author={Gambhir, Mahak and Gupta, Vishal},
  journal={Artificial Intelligence Review},
  volume={47},
  number={1},
  pages={1--66},
  year={2017},
  publisher={Springer}
}

@inproceedings{bertscore_2020,
  author       = {Tianyi Zhang and
                  Varsha Kishore and
                  Felix Wu and
                  Kilian Q. Weinberger and
                  Yoav Artzi},
  title        = {BERTScore: Evaluating Text Generation with {BERT}},
  booktitle    = {8th International Conference on Learning Representations, {ICLR} 2020,
                  Addis Ababa, Ethiopia, April 26-30, 2020},
  publisher    = {OpenReview.net},
  year         = {2020},
  url          = {https://openreview.net/forum?id=SkeHuCVFDr},
  bibsource    = {dblp computer science bibliography, https://dblp.org}
}

@misc{lewis_bart_2019,
	title = {{BART}: {Denoising} {Sequence}-to-{Sequence} {Pre}-training for {Natural} {Language} {Generation}, {Translation}, and {Comprehension}},
	shorttitle = {{BART}},
	url = {http://arxiv.org/abs/1910.13461},
	urldate = {2023-10-28},
	publisher = {arXiv},
	author = {Lewis, Mike and Liu, Yinhan and Goyal, Naman and Ghazvininejad, Marjan and Mohamed, Abdelrahman and Levy, Omer and Stoyanov, Ves and Zettlemoyer, Luke},
	month = oct,
	year = {2019},
	note = {arXiv:1910.13461 [cs, stat]},
}

@inproceedings{zhan2022CovidET,
  title={Why Do You Feel This Way? Summarizing Triggers of Emotions in Social Media Posts},
  author={Zhan, Hongli and Sosea, Tiberiu and Caragea, Cornelia and Li, Junyi Jessy},
  booktitle={Proceedings of the 2022 Conference on Empirical Methods in Natural Language Processing},
  pages={9436--9453},
  year={2022}
}

@misc{yang_exploring_2023,
	title = {Exploring the {Limits} of {ChatGPT} for {Query} or {Aspect}-based {Text} {Summarization}},
	url = {http://arxiv.org/abs/2302.08081},
	urldate = {2023-10-22},
	publisher = {arXiv},
	author = {Yang, Xianjun and Li, Yan and Zhang, Xinlu and Chen, Haifeng and Cheng, Wei},
	month = feb,
	year = {2023},
	note = {arXiv:2302.08081 [cs]},
}

@inproceedings{zhong_qmsum_2021,
	address = {Online},
	title = {{QMSum}: {A} {New} {Benchmark} for {Query}-based {Multi}-domain {Meeting} {Summarization}},
	shorttitle = {{QMSum}},
	url = {https://aclanthology.org/2021.naacl-main.472},
	doi = {10.18653/v1/2021.naacl-main.472},
	urldate = {2024-01-08},
	booktitle = {Proceedings of the 2021 {Conference} of the {North} {American} {Chapter} of the {Association} for {Computational} {Linguistics}: {Human} {Language} {Technologies}},
	publisher = {Association for Computational Linguistics},
	author = {Zhong, Ming and Yin, Da and Yu, Tao and Zaidi, Ahmad and Mutuma, Mutethia and Jha, Rahul and Awadallah, Ahmed Hassan and Celikyilmaz, Asli and Liu, Yang and Qiu, Xipeng and Radev, Dragomir},
	editor = {Toutanova, Kristina and Rumshisky, Anna and Zettlemoyer, Luke and Hakkani-Tur, Dilek and Beltagy, Iz and Bethard, Steven and Cotterell, Ryan and Chakraborty, Tanmoy and Zhou, Yichao},
	month = jun,
	year = {2021},
	pages = {5905--5921},
}

@inproceedings{wang_squality_2022,
	address = {Abu Dhabi, United Arab Emirates},
	title = {{SQuALITY}: {Building} a {Long}-{Document} {Summarization} {Dataset} the {Hard} {Way}},
	shorttitle = {{SQuALITY}},
	url = {https://aclanthology.org/2022.emnlp-main.75},
	doi = {10.18653/v1/2022.emnlp-main.75},
	urldate = {2024-01-08},
	booktitle = {Proceedings of the 2022 {Conference} on {Empirical} {Methods} in {Natural} {Language} {Processing}},
	publisher = {Association for Computational Linguistics},
	author = {Wang, Alex and Pang, Richard Yuanzhe and Chen, Angelica and Phang, Jason and Bowman, Samuel R.},
	editor = {Goldberg, Yoav and Kozareva, Zornitsa and Zhang, Yue},
	month = dec,
	year = {2022},
	pages = {1139--1156},
}

@misc{yang_oasum_2023,
	title = {{OASum}: {Large}-{Scale} {Open} {Domain} {Aspect}-based {Summarization}},
	shorttitle = {{OASum}},
	url = {http://arxiv.org/abs/2212.09233},
	language = {en},
	urldate = {2024-01-16},
	publisher = {arXiv},
	author = {Yang, Xianjun and Song, Kaiqiang and Cho, Sangwoo and Wang, Xiaoyang and Pan, Xiaoman and Petzold, Linda and Yu, Dong},
	month = may,
	year = {2023},
	note = {arXiv:2212.09233 [cs]},
}

@misc{amar_openasp_2023,
	title = {{OpenAsp}: {A} {Benchmark} for {Multi}-document {Open} {Aspect}-based {Summarization}},
	shorttitle = {{OpenAsp}},
	url = {http://arxiv.org/abs/2312.04440},
	urldate = {2024-01-18},
	publisher = {arXiv},
	author = {Amar, Shmuel and Schiff, Liat and Ernst, Ori and Shefer, Asi and Shapira, Ori and Dagan, Ido},
	month = dec,
	year = {2023},
	note = {arXiv:2312.04440 [cs]},
}

@misc{tan_summarizing_any_2020,
	title = {Summarizing {Text} on {Any} {Aspects}: {A} {Knowledge}-{Informed} {Weakly}-{Supervised} {Approach}},
	shorttitle = {Summarizing {Text} on {Any} {Aspects}},
	url = {http://arxiv.org/abs/2010.06792},
	urldate = {2024-01-21},
	publisher = {arXiv},
	author = {Tan, Bowen and Qin, Lianhui and Xing, Eric P. and Hu, Zhiting},
	month = oct,
	year = {2020},
	note = {arXiv:2010.06792 [cs]},
}

@inproceedings{vig2022exploringNeuQF,
  title={Exploring Neural Models for Query-Focused Summarization},
  author={Vig, Jesse and Fabbri, Alexander Richard and Kry{\'s}ci{\'n}ski, Wojciech and Wu, Chien-Sheng and Liu, Wenhao},
  booktitle={Findings of the Association for Computational Linguistics: NAACL 2022},
  pages={1455--1468},
  year={2022}
}

@inproceedings{pagnoni2023socratic,
  title={Socratic Pretraining: Question-Driven Pretraining for Controllable Summarization},
  author={Pagnoni, Artidoro and Fabbri, Alex and Kry{\'s}ci{\'n}ski, Wojciech and Wu, Chien-Sheng},
  booktitle={Proceedings of the 61st Annual Meeting of the Association for Computational Linguistics (Volume 1: Long Papers)},
  pages={12737--12755},
  year={2023}
}

@article{Sotudeh2023QontSumOC,
  title={QontSum: On Contrasting Salient Content for Query-focused Summarization},
  author={Sajad Sotudeh and Nazli Goharian},
  journal={ArXiv},
  year={2023},
  volume={abs/2307.07586},
  url={https://api.semanticscholar.org/CorpusID:259937277}
}

@Misc{peft,
  title =        {PEFT: State-of-the-art Parameter-Efficient Fine-Tuning methods},
  author =       {Sourab Mangrulkar and Sylvain Gugger and Lysandre Debut and Younes Belkada and Sayak Paul and Benjamin Bossan},
  howpublished = {\url{https://github.com/huggingface/peft}},
  year =         {2022}
}

@misc{touvron2023llama,
      title={LLaMA: Open and Efficient Foundation Language Models}, 
      author={Hugo Touvron and Thibaut Lavril and Gautier Izacard and Xavier Martinet and Marie-Anne Lachaux and Timothée Lacroix and Baptiste Rozière and Naman Goyal and Eric Hambro and Faisal Azhar and Aurelien Rodriguez and Armand Joulin and Edouard Grave and Guillaume Lample},
      year={2023},
      eprint={2302.13971},
      archivePrefix={arXiv},
      primaryClass={cs.CL}
}

@article{ouyang_instructGPT_2022,
  title={Training language models to follow instructions with human feedback},
  author={Ouyang, Long and Wu, Jeffrey and Jiang, Xu and Almeida, Diogo and Wainwright, Carroll and Mishkin, Pamela and Zhang, Chong and Agarwal, Sandhini and Slama, Katarina and Ray, Alex and others},
  journal={Advances in neural information processing systems},
  volume={35},
  pages={27730--27744},
  year={2022}
}

@inproceedings{
he2022parallel_adapters,
title={Towards a Unified View of Parameter-Efficient Transfer Learning},
author={Junxian He and Chunting Zhou and Xuezhe Ma and Taylor Berg-Kirkpatrick and Graham Neubig},
booktitle={International Conference on Learning Representations},
year={2022},
url={https://openreview.net/forum?id=0RDcd5Axok}
}

@inproceedings{lester2021prompt-tuning,
  title={The Power of Scale for Parameter-Efficient Prompt Tuning},
  author={Lester, Brian and Al-Rfou, Rami and Constant, Noah},
  booktitle={Proceedings of the 2021 Conference on Empirical Methods in Natural Language Processing},
  pages={3045--3059},
  year={2021}
}

@inproceedings{hu2021lora,
  title={LoRA: Low-Rank Adaptation of Large Language Models},
  author={Hu, Edward J and Wallis, Phillip and Allen-Zhu, Zeyuan and Li, Yuanzhi and Wang, Shean and Wang, Lu and Chen, Weizhu and others},
  booktitle={International Conference on Learning Representations},
  year={2021}
}

@misc{zhang_llama-adapter_2023,
	title = {{LLaMA}-{Adapter}: {Efficient} {Fine}-tuning of {Language} {Models} with {Zero}-init {Attention}},
	shorttitle = {{LLaMA}-{Adapter}},
	url = {http://arxiv.org/abs/2303.16199},
	urldate = {2024-03-10},
	publisher = {arXiv},
	author = {Zhang, Renrui and Han, Jiaming and Liu, Chris and Gao, Peng and Zhou, Aojun and Hu, Xiangfei and Yan, Shilin and Lu, Pan and Li, Hongsheng and Qiao, Yu},
	month = jun,
	year = {2023},
	note = {arXiv:2303.16199 [cs]},
}

@misc{hu_llm-adapters_2023,
	title = {{LLM}-{Adapters}: {An} {Adapter} {Family} for {Parameter}-{Efficient} {Fine}-{Tuning} of {Large} {Language} {Models}},
	shorttitle = {{LLM}-{Adapters}},
	url = {http://arxiv.org/abs/2304.01933},
	urldate = {2024-03-10},
	publisher = {arXiv},
	author = {Hu, Zhiqiang and Wang, Lei and Lan, Yihuai and Xu, Wanyu and Lim, Ee-Peng and Bing, Lidong and Xu, Xing and Poria, Soujanya and Lee, Roy Ka-Wei},
	month = oct,
	year = {2023},
	note = {arXiv:2304.01933 [cs]},
}

@inproceedings{ha2016hypernetworks,
  title={HyperNetworks},
  author={Ha, David and Dai, Andrew M and Le, Quoc V},
  booktitle={International Conference on Learning Representations},
  year={2016}
}

@article{munkhdalai2024infinitransformer,
  title={Leave no context behind: Efficient infinite context transformers with infini-attention},
  author={Munkhdalai, Tsendsuren and Faruqui, Manaal and Gopal, Siddharth},
  journal={arXiv preprint arXiv:2404.07143},
  year={2024}
}

@misc{ivison_hyperdecoders_2022,
	title = {Hyperdecoders: {Instance}-specific decoders for multi-task {NLP}},
	shorttitle = {Hyperdecoders},
	url = {http://arxiv.org/abs/2203.08304},
	urldate = {2024-02-16},
	publisher = {arXiv},
	author = {Ivison, Hamish and Peters, Matthew E.},
	month = oct,
	year = {2022},
	note = {arXiv:2203.08304 [cs]},
}

@misc{he_hyperprompt_2022,
	title = {{HyperPrompt}: {Prompt}-based {Task}-{Conditioning} of {Transformers}},
	shorttitle = {{HyperPrompt}},
	url = {http://arxiv.org/abs/2203.00759},
	urldate = {2024-03-25},
	publisher = {arXiv},
	author = {He, Yun and Zheng, Huaixiu Steven and Tay, Yi and Gupta, Jai and Du, Yu and Aribandi, Vamsi and Zhao, Zhe and Li, YaGuang and Chen, Zhao and Metzler, Donald and Cheng, Heng-Tze and Chi, Ed H.},
	month = jun,
	year = {2022},
	note = {arXiv:2203.00759 [cs]},
}

@misc{zhao_hypermoe_2024,
	title = {{HyperMoE}: {Paying} {Attention} to {Unselected} {Experts} in {Mixture} of {Experts} via {Dynamic} {Transfer}},
	shorttitle = {{HyperMoE}},
	url = {http://arxiv.org/abs/2402.12656},
	urldate = {2024-03-03},
	publisher = {arXiv},
	author = {Zhao, Hao and Qiu, Zihan and Wu, Huijia and Wang, Zili and He, Zhaofeng and Fu, Jie},
	month = feb,
	year = {2024},
	note = {arXiv:2402.12656 [cs]},
}

@inproceedings{katharopoulos2020transformersarernn,
  title={Transformers are rnns: Fast autoregressive transformers with linear attention},
  author={Katharopoulos, Angelos and Vyas, Apoorv and Pappas, Nikolaos and Fleuret, Fran{\c{c}}ois},
  booktitle={International conference on machine learning},
  pages={5156--5165},
  year={2020},
  organization={PMLR}
}

@article{clevert2015elus,
  title={Fast and accurate deep network learning by exponential linear units (elus)},
  author={Clevert, Djork-Arn{\'e} and Unterthiner, Thomas and Hochreiter, Sepp},
  journal={arXiv preprint arXiv:1511.07289},
  year={2015}
}

@inproceedings{schlag2020associative,
  title={Learning Associative Inference Using Fast Weight Memory},
  author={Schlag, Imanol and Munkhdalai, Tsendsuren and Schmidhuber, J{\"u}rgen},
  booktitle={International Conference on Learning Representations},
  year={2020}
}

@article{beltagy2020longformer,
  title={Longformer: The long-document transformer},
  author={Beltagy, Iz and Peters, Matthew E and Cohan, Arman},
  journal={arXiv preprint arXiv:2004.05150},
  year={2020}
}

@inproceedings{zhu2020HMNet,
  title={A Hierarchical Network for Abstractive Meeting Summarization with Cross-Domain Pretraining},
  author={Zhu, Chenguang and Xu, Ruochen and Zeng, Michael and Huang, Xuedong},
  booktitle={Findings of the Association for Computational Linguistics: EMNLP 2020},
  pages={194--203},
  year={2020}
}

@article{bertsch2024unlimiformer,
  title={Unlimiformer: Long-range transformers with unlimited length input},
  author={Bertsch, Amanda and Alon, Uri and Neubig, Graham and Gormley, Matthew},
  journal={Advances in Neural Information Processing Systems},
  volume={36},
  year={2024}
}

@article{achiam2023gpt,
  title={Gpt-4 technical report},
  author={Achiam, Josh and Adler, Steven and Agarwal, Sandhini and Ahmad, Lama and Akkaya, Ilge and Aleman, Florencia Leoni and Almeida, Diogo and Altenschmidt, Janko and Altman, Sam and Anadkat, Shyamal and others},
  journal={arXiv preprint arXiv:2303.08774},
  year={2023}
}

@article{yang2025power,
  title={The power of ChatGPT in processing text: Evidence from analysis and prediction in the exchange rate markets},
  author={Yang, Kun and Deng, Ruxin and Wei, Yunjie and Wang, Shouyang},
  journal={Financial Innovation},
  volume={11},
  number={1},
  pages={118},
  year={2025},
  publisher={Springer}
}

@article{cary2024herding,
  title={Herding and investor sentiment after the cryptocurrency crash: evidence from Twitter and natural language processing},
  author={Cary, Michael},
  journal={Financial Innovation},
  volume={10},
  number={1},
  pages={142},
  year={2024},
  publisher={Springer}
}

@article{yang2025qwen3,
  title={Qwen3 technical report},
  author={Yang, An and Li, Anfeng and Yang, Baosong and Zhang, Beichen and Hui, Binyuan and Zheng, Bo and Yu, Bowen and Gao, Chang and Huang, Chengen and Lv, Chenxu and others},
  journal={arXiv preprint arXiv:2505.09388},
  year={2025}
}








\end{document}